\documentclass{article}
\usepackage{tikz} % Package for drawing
\usepackage{amsmath}
\allowdisplaybreaks
\usepackage{wasysym}
\usepackage{amssymb}
\usepackage{hyperref}
\usepackage{listings}
\usepackage{dsfont}
\usepackage{graphicx}
\usepackage{subcaption}
\usepackage{multirow}
\usepackage[round]{natbib} % enables pretty citations
\usepackage{enumitem}% enables pretty lists
\usepackage{authblk}
\usepackage[margin=1in, includefoot]{geometry}
\usepackage{algorithm,algpseudocode}
\usepackage{bm}
\usepackage{tabu} 
\usepackage{multirow}
\usepackage{comment}
\usepackage{xspace}
\usepackage{nicefrac}
\usepackage{lipsum}
\usepackage{lineno}
\usepackage[T3,T1]{fontenc}
\DeclareSymbolFont{tipa}{T3}{cmr}{m}{n}
\DeclareMathAccent{\invbreve}{\mathalpha}{tipa}{16}
\usepackage{cases}
\usepackage{array}
\newcolumntype{P}[1]{>{\centering\arraybackslash}p{#1}}

\usepackage[flushleft]{threeparttable}
\usepackage[toc,page]{appendix}
\usepackage{graphicx} % Required for inserting images

\title{MM-algorithms for traditional and convex NMF with Tweedie and Negative Binomial cost functions and empirical evaluation}
\author{Elisabeth Sommer James, Asger Hobolth, Marta Pelizzola}
\date{  %
    {\small Department of Mathematics, Aarhus University, Denmark \\ \{elijam, asger, marta\}@math.au.dk}\\%
    [2ex]%
    \today
}

\begin{document}

\maketitle

\section*{Abstract} \label{abstract}
Non-negative matrix factorisation (NMF) is a widely used tool for unsupervised learning and feature extraction, with applications ranging from genomics to text analysis and signal processing. Standard formulations of NMF are typically derived under Gaussian or Poisson noise assumptions, which may be inadequate for data exhibiting overdispersion or other complex mean–variance relationships. In this paper, we develop a unified framework for both traditional and convex NMF under a broad class of distributional assumptions, including Negative Binomial and Tweedie models, where the connection between the Tweedie and the $\beta$-divergence is also highlighted. Using a Majorize–Minimisation approach, we derive multiplicative update rules for all considered models, and novel updates for convex NMF with Poisson and Negative Binomial cost functions. We provide a unified implementation of all considered models, including the first implementations of several convex NMF models. Empirical evaluations on mutational and word count data demonstrate that the choice of noise model critically affects model fit and feature recovery, and that convex NMF can provide an efficient and robust alternative to traditional NMF in scenarios where the number of classes is large. The code for our proposed updates is available in the R package \texttt{nmfgenr} and can be found at \url{https://github.com/MartaPelizzola/nmfgenr}. \\
\bigskip
\textbf{Keywords:} convex NMF, MM-algorithms, model choice, NMF

\section{Introduction} \label{introduction}
%% nmf
Non-negative matrix factorisation (NMF) is a widely used unsupervised learning technique, originally introduced by \cite{Lee1999} to extract parts-based representations from image and text data.
Since its introduction, NMF and its many extensions have found applications across a broad range of domains such as the analysis of cancer mutations \citep{alexandrov2013signatures} and gene expression data \citep{Kim2007} in genetics, speech \citep{Mohammadiha2013} and signal \citep{Leplat2020} processing and data imputation \citep{Qian2024}.

NMF approximates a data matrix $V \in \mathbb{R}_+^{N\times M}$ as the product of two non-negative matrices $W \in \mathbb{R}_+^{N\times K}$ and $H \in \mathbb{R}_+^{K\times M}$ such that
$$
V \approx WH,
$$ where $H$ is commonly described as the feature matrix as each of the $K$ rows of $H$ can be interpreted as a feature, and $W$ the matrix containing the corresponding weights of each feature. The rank  $K$ is typically chosen to be much smaller than $N$ and $M$. 

%% convex nmf
Alternative forms of NMF have been proposed in the literature, in particular for enhancing sparsity. \cite{PascualMontano2006, Kim2007, Lal2021} introduced  several modifications of traditional NMF with sparseness constraints to the factors and \cite{Ding2010} developed a framework for convex NMF. In this model features are linear combinations of the data matrix, $V$. The factorisation is given by $$V^T \approx V^TED,$$ where the superscript $T$ denotes the matrix transpose. The matrices $E, D^T \in \mathbb{R}_+^{N \times K}$ are estimated.  The weights matrix of the traditional NMF is equivalent to $D^T$ and the features matrix is equivalent to $(V^TE)^T$. \cite{Egendal2025relation} has shown that this convex NMF model is equivalent to a shallow linear autoencoder where the bias terms are set to zero. In the autoencoder formulation of this model, $E$ corresponds to the encoder matrix and $D$ to the decoder matrix. 

%% distributional assumption / cost function choice
In both standard and convex NMF, the choice of cost function is crucial, as it implicitly assumes a corresponding distribution of the data. The formulations of \cite{Lee1999} minimise either least squares error or generalized Kullback–Leibler divergence, corresponding to Gaussian and Poisson models. In practice, however, many applications exhibit overdispersion, heavy tails, or other departures from these assumptions, thus requiring different distributions for the underlying NMF model. Assessing whether the data are well modelled by the assumed distribution is highly important as incorrect assumptions can distort the factorisation since the estimates of $W$ and $H$ depend directly on the chosen cost function.

%% available example from literature 
For this reason, other versions of NMF and convex NMF have been proposed in the literature beyond the classical Gaussian and Poisson versions. For instance, the Negative Binomial distribution has been employed to account for overdispersion relative to the Poisson model \citep{Gouvert2020}, which is particularly relevant in applications such as cancer mutational counts data \citep{Pelizzola2023} or single-cell RNA-seq data \citep{Matsumoto2019} where the variance of the data exceeds the mean. Lastly, the $\beta$-divergence NMF \citep{fevotte2011} allows to fit a wide range of noise models, including heavy-tailed noise common in image processing \citep{Lam2008} and text mining \citep{Kalcheva2020}. This measure is equivalent to minimising the negative log-likelihood of the Tweedie distribution which provides an alternative flexible model for NMF. The different mean-variance relation of these models can be used to effectively perform model selection when applying NMF or convex NMF to a data set thereby improving the reliability of the resulting factorisation. 

\begin{table}[h] 
\centering
\begin{threeparttable}
\begin{tabular}{c | cc cc}
\hline
\multirow{2}{*}{Distribution} & \multicolumn{2}{c}{Traditional} & \multicolumn{2}{c}{Convex} \\ %& \multirow{2}{*}{Applications} \\
\cline{2-3} \cline{4-5}
& Algorithm & Implementation & Algorithm & Implementation  \\
\hline
Normal & \checkmark$^{\tnote{1}}$ & \checkmark$^{\tnote{4}}$ & \checkmark$^{\tnote{5}}$ & \textbf{\CheckedBox}  \\ %Discovery of overlooked differentially expressed (DE) gene regions \par \cite{Matsumoto2019}  \\
Poisson & \checkmark$^{\tnote{1}}$ & \checkmark$^{\tnote{4}}$ & \checkmark$^{\tnote{2}}$ & \textbf{\CheckedBox} \\  %Spatial transcriptomics inference \par \citep{laursen2025neighborhood} \\
Tweedie & \checkmark$^{\tnote{2}}$ & \checkmark$^{\tnote{2}}$ & \checkmark$^{\tnote{2}}$ & \textbf{\CheckedBox} \\ %Music signal decomposition \par \citep{fevotte2009nonnegative} \\
Negative Binomial & \checkmark$^{\tnote{3}}$ & \checkmark$^{\tnote{3}}$ & \textbf{\CheckedBox} & \textbf{\CheckedBox} \\ %& Inference of mutational signatures \citep{Pelizzola2023} \\
\hline
\end{tabular}
  \begin{tablenotes}
    \item[1] \cite{Lee1999}
    \item[2] \cite{fevotte2011}
    \item[3] \cite{Pelizzola2023}
    \item[4] \cite{Brunet2004}
    \item[5] \cite{Ding2010}; an alternative MM-algorithm is derived in this manuscript (Section \ref{sec:normalnmf}) with a different choice of the majorization function. 
  \end{tablenotes}
  \end{threeparttable}
\caption{Overview of traditional and convex non-negative matrix factorisation models under different distributional assumptions. For each distribution, existing estimation procedure and implementation are indicated with \checkmark; new contributions introduced in this paper are indicated with \textbf{\CheckedBox}. References for existing algorithms and implementation are provided as footnotes.}
\label{tab:overview}
\end{table}

%% contribution of paper 1 MM algorithms 
In this paper we extend the available update rules for traditional and convex NMF by deriving new multiplicative updates via the Majorize-Minimisation (MM) algorithm for Negative Binomial Convex NMF and providing ready-to-use implementations of the estimation procedures for convex NMF under all considered distributional assumptions as outlined in Table \ref{tab:overview}.  We choose MM multiplicative updates to optimise NMF and convex NMF under the different distributional assumptions as these are closed-form updates tailored to the different cost functions and we provide a computationally efficient solution to optimise NMF. A commonly used alternative way to optimise NMF is projected gradient descent which is often preferred for its simple implementation, but shows very slow convergence \citep{Hobolth2019}.  Our newly derived multiplicative updates extend the family of models available within the convex NMF framework, especially for settings with overdispersed or heavy-tailed data. The proposed implementations and corresponding R package allow to apply these models easily to real data in diverse settings. 

%% contribution of paper 2 implementation
Indeed, we provide a fast and easy to use implementation of Tweedie and Negative Binomial NMF and convex NMF updates. Using the correct noise model is essential for accurate feature extraction \citep{Pelizzola2023}, and our work enables researchers to explore models aligned with the mean–variance structure of their data by offering accessible implementations under different distributional assumptions. 

%% contribution of paper 3 convex NMF
Lastly, studying convex NMF and how it compares to traditional NMF on different applications and with various cost functions is also a timely question. The convex NMF model has been shown to be equivalent to a shallow autoencoder \citep{Egendal2025relation}. As autoencoders have become increasingly popular for feature extraction \citep{Smaragdis2017, Ozer2022, Pancotti2024}, exploring convex NMF for feature extraction improves our understanding on this more interpretable alternative to autoencoders. Furthermore, \cite{Ding2010} argued that convex NMF is better when modelling sparse data which is also important as real data become larger and thus often sparser. An overview of available and newly derived multiplicative updates, corresponding implementations and relevant applications can be found in Table \ref{tab:overview}.

%% considered applications
We compare our proposed versions of NMF and convex NMF on two different applications. First, we consider a collection of mutational counts data from 260 liver cancer patients from  \cite{Campbell2020}. Here, NMF is used to find mutational signatures, i.e.\, probability distributions representing the mutational processes operating in the different tumors, and the corresponding weights which correspond to the exposure to the mutational processes active in each patient during cancer evolution. Precise identification of mutational signatures is relevant for clinical purposes to define treatments for cancer patients \citep{Caruso2017} or develop prevention strategies \citep{Zhang2021}. The second application considers word counts extracted from posts to 20 different newsgroups from \cite{Rennie2007}. Here, we take a subset of 500 documents from three different topics (sport, religion and politics) and 6354 words and use NMF and convex NMF to find topics discussed in the considered posts. Identifying topics in large sets of text data is relevant for classification and information retrieval \citep{onan2016}. These data are typically highly sparse and illustrate the performance of our proposed methods on sparse datasets, which are common across many NMF applications. Our analyses show that the choice of the correct NMF model and cost function are essential: we find that the performance of the different models differ between applications and show that convex NMF is indeed beneficial on sparse data. 

%% structure of paper and code availability
The remaining part of the paper is organised as follows. In Section \ref{sec:methods} multiplicative update rules for traditional and convex NMF with Tweedie (Section \ref{sec:tweedie}) and Negative Binomial (Section \ref{sec:negbin}) distributions are derived. Furthermore, we provide details on computational aspects of the different procedures in Section \ref{sec:computational}. In Section \ref{sec:results} we show an application of NMF and convex NMF under the different cost functions on the cancer genomic and the text data set.  In Section \ref{sec:discussion}, we provide a discussion of the various approaches presented in this manuscript. 
We provide a fast implementation in Rcpp suitable for large data sets of all considered versions of NMF and convex NMF under different cost functions in the R package \texttt{nmfgenr} available at \url{https://github.com/MartaPelizzola/nmfgenr}. 

\section{Methods} \label{sec:methods}
We provide an overview of update rules for traditional and convex NMF with different underlying model choices and corresponding cost functions. In particular, we include the Tweedie distribution (with Poisson and Normal distributions as special cases) and the Negative Binomial distribution. In this work, updates for convex NMF under the Negative Binomial distribution are newly derived. For reproducibility and practical use, implementations of all proposed methods are provided to facilitate application to real data.

\subsection{Notation} \label{sec:notation}
In order to specify the model and corresponding choice of cost function and parameters we suggest using Kendall's notation \citep{Kendall1953}. Developed for queueing theory to describe and classify a queueing node, it provides a way of summarising the characteristics of each model in a compact way.
In the following, we denote traditional NMF \citep{Lee1999} by $\mathcal{T}$ and convex NMF \citep{Ding2010} by $\mathcal{C}$, and $K$ as the rank of the given factorisation. For the distributions specifying the assumed cost functions we use the notation in Table~\ref{tab:Kendall}.

\begin{table}[h]
\centering
\begin{tabular}{ll}
\hline
\multicolumn{1}{c}{Symbol} & \multicolumn{1}{c}{Distribution}                                 \\ \hline
$\mathcal{N}$                         & Normal distribution       \\
Po                         & Poisson distribution  \\
TW$_p$                         & Tweedie distribution with power $p$                      \\
NB$_\alpha$                         & Negative Binomial distribution with dispersion $\alpha$  \\
\hline
\end{tabular}
\caption{Abbreviations for distributions specifying the cost functions.}
\label{tab:Kendall}
\end{table}

A specific model with a given cost function can be described using Kendall's notation with the following structure: \\
\centerline{NMF/[model for NMF]/distribution$_{\text{parameter}}$/rank.} 
\vspace{0.013889in} \\
For example NMF/$\mathcal{T}$/TW$_{2}$/3 would be traditional NMF with rank 3 and the Tweedie distribution with power $p = 2$ as cost function. To refer to an NMF model without a specific distributional assumption we use NMF/$\mathcal{T}$ for traditional NMF and NMF/$\mathcal{C}$ for convex NMF. We omit the rank in our notation when referring to models with no fixed rank. 

\subsection{Tweedie distribution}\label{sec:tweedie}
In this section we describe the update rules for NMF/$\mathcal{T}$/TW$_{p}$ and NMF/$\mathcal{C}$/TW$_{p}$. Furthermore, we note that the Normal and Poisson traditional and convex NMF are special cases of the Tweedie (where TW$_0$ corresponds to the Normal and TW$_1$ to the Poisson) considered here and we summarise those updates too.  

\begin{figure}[h!]
    \centering
    \includegraphics[width=\linewidth]{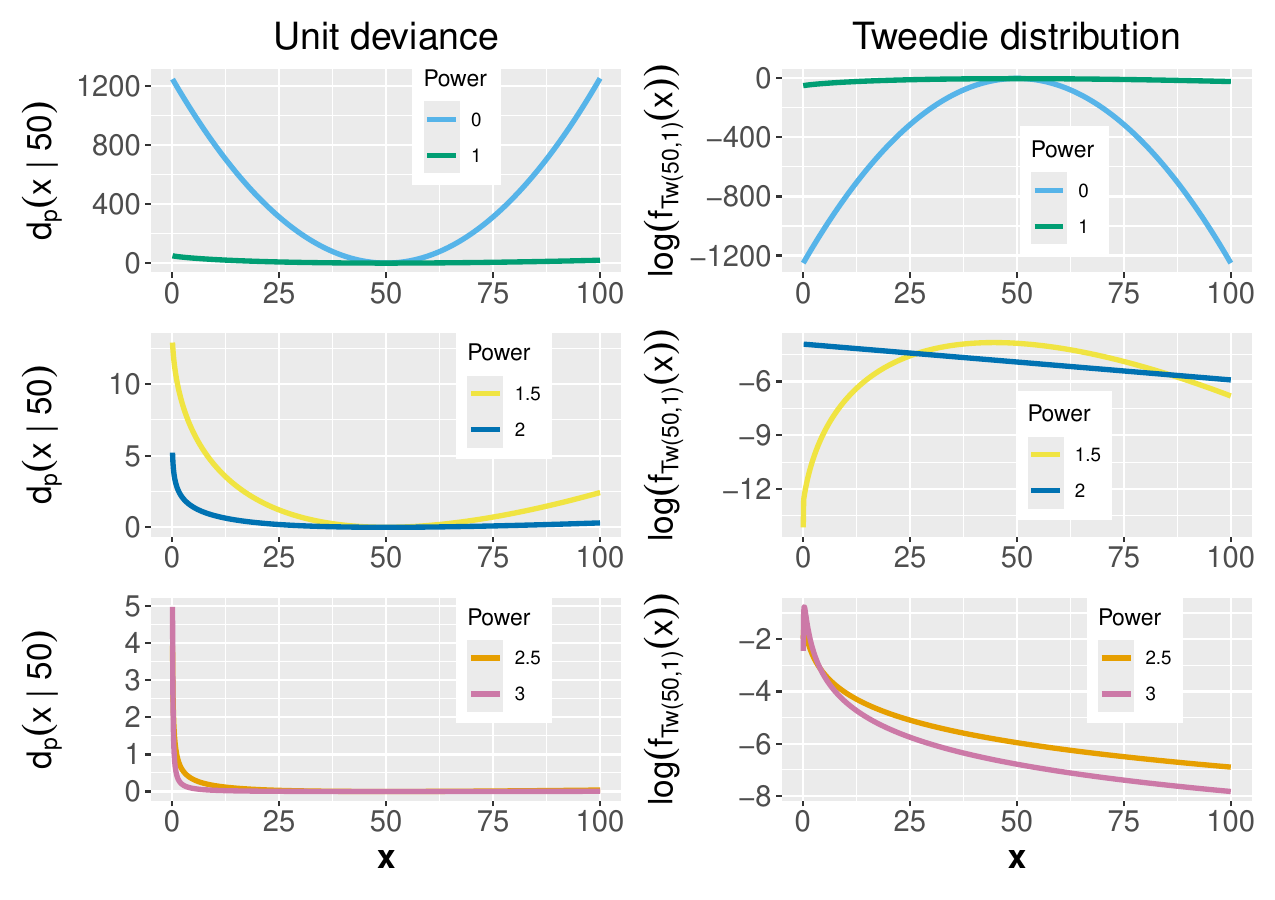}
    \caption{Unit deviances and the corresponding Tweedie distributions with parameters $(\mu, \sigma^2) = (50,1)$ for $p \in \{0,1\}$, $p \in \{1.5,2\}$, and $p \in \{2.5,3\}$.}
    \label{fig:tweediedivpdf}
\end{figure}

Let $X$ be Tweedie distributed with power parameter $p$, mean $\mu$ and dispersion parameter $\sigma^2 > 0$ \citep{jorgensen1997theory}. The support and parameter space are defined in Table \ref{table:tweediedist}. We write $$X \sim \text{Tw}_p(\mu, \sigma^2).$$ 
The Tweedie distribution is undefined for $p \in (0,1)$ and, whilst the distribution is defined for $p < 0$, we do not include it here as it is generally considered an unrealistic modelling assumption. This means we consider $p=0$ or $p\geq 1$. The probability density function can be written in the form 
\begin{align} \label{eq:tweediedensity}
    f_{\text{Tw}_p(\mu, \sigma^2)}(x) = c(x, \sigma^2)\exp\left(-d_p(x;\mu)/2\sigma^2\right),
\end{align}
where $d_p(x;\mu)$ is the unit deviance, defined as 
 \begin{numcases}{d_p(x;\mu) =}
    2 \left[ \frac{x^{2-p}}{(1-p)(2-p)} - \frac{x\mu^{1-p}}{1-p} + \frac{\mu^{2-p}}{2-p} \right] & $p \in \mathbb{R}_0^+\setminus \{1,2 \}$ \label{eq:unitdevpgen} \\ 
    2\left[x\log(\frac{x}{\mu}) -x + \mu\right] & $p = 1$\label{eq:unitdevp1} \\ 
    2\left[\frac{x}{\mu} - \log (\frac{x}{\mu}) - 1\right] & $p=2$. \label{eq:unitdevp2}
\end{numcases}
The parameter $p$ refers to the variance of $X$, which is linked to the mean by a power law: $$\mathbb{E}[X] = \mu \text{ and } \text{Var}X = \sigma^2 \mu^p.$$
The Tweedie distribution is symmetric for $p = 0$, and otherwise asymmetric as shown in Figure \ref{fig:tweediedivpdf}. Figure \ref{fig:tweediedivpdf} also shows that the Tweedie distribution is suitable to model heavy tailed data for $p > 2$ (lower panel). 
\begin{table}[h!]
    \centering
    {\def\arraystretch{1.5}
    \makebox[\textwidth]{    \begin{tabular}{c c | c c | c | c | c}
       Power & Distribution &  Unit deviance & Divergence & Normalising & Support & Parameter\\[-2ex]
    & & $d_p(x;\mu)$ & &factor, $c(x,\sigma^2)$ & of $x$ & space of $\mu$ \\ \hline
    $p = 0$ & Normal  & $(x-\mu)^2$  & Euclidean& $1/{\sqrt{2\pi\sigma^2}}$&$\mathbb{R }$&$\mathbb{R }$ 
    \\
    $p = 1$ & Poisson  & $2(x\log(\frac{x}{\mu}) - (x-\mu))$ & Kullback- & $\frac{x^x}{x!e^{x}} $& $\mathbb{N}_0$&$\mathbb{R}_+$ \\[-2ex]
    &&&Leibler&&& \\
    $1 < p < 2$ & Compound & $d_p(x;\mu)$ & - & - & $[0,\infty)$ & $\mathbb{R}_+$ \\ [-2ex]
    &Poisson&&&&& \\
    $p = 2$ & Gamma  &  $2(\log(\frac{\mu}{x}) + \frac{x}{\mu} - 1)$ &  Itakura-Saito & $\frac{\sigma^{-2\frac{1}{\sigma^2}}e^{-\frac{1}{\sigma^2}}}{x\Gamma(\frac{1}{\sigma^2})}$ & $\mathbb{R}_+$&$\mathbb{R}_+$\\
    $p \in (2,3) $ & Positive & $d_p(x;\mu)$ & - & - & $\mathbb{R}_+$ & $\mathbb{R}_+$ \\[-2ex]
    &stable&&&&& \\
    $p = 3$ & Inverse&   $\left( \frac{\sqrt{x}}{\mu} - \frac{1}{\sqrt{x}}\right)^2$  & Transformed &$\sqrt{\frac{1}{2\pi x^3\sigma^2}}$ & $\mathbb{R}_+$& $\mathbb{R}_+$\\[-2ex]
    &Gaussian&&Euclidean&&& \\
    $p \in (3,\infty)$ & Positive & $d_p(x;\mu)$ & - & - & $\mathbb{R}_+$ & $\mathbb{R}_+$ \\[-2ex]
    &stable&&&&& \\
    \end{tabular}}}
    \caption{Overview of the Tweedie distribution for $p\geq 0$ (adapted from \cite{jorgensen1997theory}).}
    \label{table:tweediedist}
\end{table}

The variance power law has been shown to be applicable in numerous contexts, such as in ecological populations \citep{kendal2002spatial}, where it is known as Taylor's power law, infectious disease modelling \citep{enki2017taylor} and human cancer metastases \citep{kendal2000characterization}. This distribution also allows to model underdispersion in applications if $\sigma^2 < 1$ and $p < 0$.
For some values of $p$, the Tweedie distribution corresponds to known distributions (see Table \ref{table:tweediedist}) but typically the normalising factor is analytically intractable. Note that for the compound Poisson distribution it is possible to derive the normalising factor, however we refer to \cite{jorgensen1997theory} for this derivation and do not report it here for simplicity of notation.

For a non-negative matrix $V \in \mathbb{R}_+^{N \times M}$, the model assumed by NMF is given by $V_{nm} \sim \text{Tw}_p(\mu_{nm}, \sigma^2) $ with $ \mu_{nm} = (WH)_{nm} $ for traditional NMF. The analogous model for convex NMF would take $ \mu_{nm} = (V^TED)_{nm} $, however this is a function of the random matrix $V$. To address this issue, we use an approach which is commonly used in the neural network literature (e.g.\ \cite{Pancotti2024}) and use the cost function directly to define NMF/$\mathcal{C}/$Tw$_p$. Then the log-likelihood function for traditional NMF can be obtained from equation \eqref{eq:tweediedensity} as
\begin{align} \label{eq:logliktweedieNMF}
  \ell(W,H,\sigma^2;V,p) &= \sum_{m= 1}^M\sum_{n=1}^N\left( \log(c(V_{nm}, \sigma^2)) -\frac{1}{2\sigma^2}d_p(V_{nm};(WH)_{nm})\right) \\ \nonumber
  &= C(\sigma^2,V) -\frac{1}{2\sigma^2}\sum_{m= 1}^M\sum_{n=1}^N d_p(V_{nm};(WH)_{nm})
\end{align}
where $C$ is a constant with respect to $W,\ H$ and $p$. For convex NMF the analogous function is
\begin{align} \label{eq:logliktweedieCNMF}
  \ell(E,D,\sigma^2;V,p) &= \sum_{m= 1}^M\sum_{n=1}^N\left( \log(c(V^T_{mn}, \sigma^2)) -\frac{1}{2\sigma^2}d_p(V^T_{mn};(V^TED)_{mn})\right) \\ \nonumber
  &= C(\sigma^2,V) -\frac{1}{2\sigma^2}\sum_{m= 1}^M\sum_{n=1}^N d_p(V^T_{mn};(V^TED)_{mn})
\end{align}
where $C$ is again a constant with respect to $E,\ D$ and $p$. Note that the minimisation problem does not involve the variance parameter $\sigma^2$.
Minimising the unit deviance between $V_{nm}$ and $ (WH)_{nm} $ and $ (V^TED)^T_{nm} $ respectively is equivalent to maximising the log likelihood of the Tweedie model defined in equations \eqref{eq:logliktweedieNMF} and \eqref{eq:logliktweedieCNMF}, and thus these forms of NMF have assumed the underlying model is a Tweedie model.

Notice that taking $p = 2-\beta $ gives the $\beta$-divergence \citep{yilmaz2012alpha}. Updates for NMF/$\mathcal{T}$/TW$_{p}$ and NMF/$\mathcal{C}$/TW$_{p}$  which assume an underlying Tweedie distribution have been derived by \cite{fevotte2009nonnegative} under the name $\beta$-divergence NMF. In general the unit deviance is not convex, and thus to define a majorizing function, it is decomposed into convex, concave and constant parts which are each majorized separately. Then a Majorize-Minimisation (MM) algorithm can be derived.

For NMF/$\mathcal{T}$/TW$_{p}$ these updates are:
\begin{align} \label{eq:NMF-TW-H-W}
H^{(t+1)} = H^t \cdot \left( \frac{W^T \left[\frac{V}{(WH) ^{\cdot p}}\right] }{W^T\left[(WH) ^{ \cdot (1 -p)}\right]} \right)^{\cdot \gamma(p)} \text{and }
W^{(t+1)} = W^t \cdot \left( \frac{\left[\frac{V}{(WH) ^{\cdot p}} \right]H^T}{\left[(WH)^{\cdot (1 - p)}\right]H^T} \right)^{\cdot \gamma(p)}, 
\end{align}
and for NMF/$\mathcal{C}$/TW$_{p}$ they are: 
\begin{align} \label{eq:CNMF-TW-E-D}
E^{(t+1)} = E^t \cdot \left( \frac{V [\frac{V^T}{(V^TE^tD) ^{\cdot p}}] D^T}{V[(V^TE^tD) ^{ \cdot (1 -p)}]D^T} \right)^{\cdot \gamma(p)}
\text{and }
D^{(t+1)} = D^t \cdot \left( \frac{(V^TE)^T[ \frac{V^T}{(V^TED^t) ^{\cdot p}}] }{(V^TE)^T\big[(V^TED^t)^{\cdot (1 - p)}\big]} \right) ^{\cdot \gamma(p)}
\end{align}
with the $\gamma$ function defined piecewise as 

$$
\gamma(p) = \begin{cases}
    1 & p = 0 \\
    \frac{1}{p} & p \geq 1.
\end{cases}
$$
Here division is always element wise, indices and multiplication denoted with `$ \cdot$' are element wise, otherwise multiplication is matrix wise. In applications, we propose to find an estimate of $p$ by profile likelihood as described in Appendix \ref{app:A}. Packages for estimating the likelihood and log likelihood of the Tweedie distribution are widely available for $p \leq 2$. For larger values of $p$, implementation is not widely available, barring the values of $p$ corresponding to known distributions, i.e. $p=3$ being the inverse Gaussian distribution. As can be seen in Figure \ref{fig:tweediedivpdf}, $p = 3$ can model strongly skewed data and thus it is reasonable in applications to only consider $p \leq 3$.
\subsubsection{Special cases: Normal and Poisson distribution} \label{sec:normalnmf}
As shown in Table \ref{table:tweediedist}, setting $p=0$ in the Tweedie distribution we recover the Normal distribution where 
\begin{align*} 
V_{nm} \sim \mathcal{N}(\mu_{nm}, \sigma^2)
\end{align*}
with $\mu_{nm} = (WH)_{nm}$ for traditional NMF and $\mu_{nm} = (V^TED)^T_{nm}$ for convex NMF. 
This implies that setting $p=0$ in the update rules in equations \eqref{eq:NMF-TW-H-W} and \eqref{eq:CNMF-TW-E-D} we recover the NMF and convex NMF updates under the normal distribution in Table \ref{tab:normalpoissonupdates}. The updates for NMF/$\mathcal{T}$/$\mathcal{N}$ have been previously derived in \cite{Lee1999}. \cite{Ding2010} also proposed updates for the same model which differ from the one showed here by a square root due to an alternative version of the MM-algorithm used. We note that MM-algorithms are not unique and different choices of the majorizing function can lead to different update rules \citep{Lange2000}. On a similar note, the implementation for NMF/$\mathcal{C}$/$\mathcal{N}$  used in this manuscript does not correspond to the one reported in Table \ref{tab:overview} and has been derived here with an alternative majorizing function. The corresponding updates are given in Table \ref{tab:normalpoissonupdates}.

Setting $(p, \sigma^2) = (1,1)$ in the Tweedie distribution corresponds to the Poisson distribution and thus we can also obtain the NMF and convex NMF updates under the Poisson distribution from equations \eqref{eq:NMF-TW-H-W} and \eqref{eq:CNMF-TW-E-D}  by setting $p=1$. Here 
\begin{align*} 
V_{nm} \sim \text{Po}(\mu_{nm})
\end{align*}
with $\mu_{nm} = (WH)_{nm}$ for traditional NMF and $\mu_{nm} = (V^TED)^T_{nm}$ for convex NMF. 

The updates introduced in this section are summarised in Table \ref{tab:normalpoissonupdates} and are well known in the literature \citep{Lee1999, Ding2010}. We note that the updates for traditional and convex NMF are very similar in structure. In particular, making the equivalence of $W = D^T$ and $WH = (V^TED)^T$, the updates for the weights matrices are the same. Due to the differing structure of the features matrices, the updates for $H$ and $E$ cannot be shown to be exactly equivalent, however by taking the same equivalences as before their structures are notably similar.

\begin{table}[h!]
\centering
\begin{tabular}{l|ll}
\cline{2-3} \\[-1.5ex]
    & Normal distribution & Poisson distribution \\[1ex] \cline{2-3} \\[-2ex] \cline{2-3} \\[-1.5ex]
    Update & NMF/$\mathcal{T}$/$\mathcal{N}$    & NMF/$\mathcal{T}$/Po     \\ [1ex] \hline 
$H^{(t+1)}$ &   $H^t \cdot \left( \frac{W^TV }{W^T WH } \right)  $  &  $ H^t \cdot \left( \frac{W^T \left[\frac{V}{WH }\right] }{W^T\ J_{N,M}} \right)$                                      \\ 
$W^{(t+1)}$ &   $W^t \cdot \left( \frac{VH^T}{WHH^T} \right)$   &    $W^t \cdot \left( \frac{\left[\frac{V}{(WH)} \right]H^T}{J_{N,M}\ H^T} \right)$                                      \\\hline \\[-2ex] \cline{2-3} \\[-1.5ex]
    & NMF/$\mathcal{C}$/$\mathcal{N}$    & NMF/$\mathcal{C}$/Po     \\ [1ex] \hline 
$E^{(t+1)}$ &  $ E^t \cdot \left( \frac{V V^T D^T}{V(V^TE^tD)D^T} \right)$   &  $E^t \cdot \left( \frac{V \left[\frac{V^T}{(V^TE^tD)}\right] D^T}{V \ J_{M,N} \ D^T} \right)$    \\
$D^{(t+1)} $ & $D^t \cdot \left( \frac{(V^TE)^TV^T }{(V^TE)^T(V^TED^t)} \right)$ &  $D^t \cdot \left( \frac{(V^TE)^T \left[ \frac{V^T}{V^TED^t }\right] }{(V^TE)^T\ J_{M,N}} \right)$                                        \\ \hline
\end{tabular}
\caption{Updates for NMF and convex NMF with the Normal and Poisson distributions. $J_{M,N}$ denotes an $M \times N$ matrix of ones.}
\label{tab:normalpoissonupdates}
\end{table}
\subsection{Negative Binomial distribution} \label{sec:negbin}
In this section we describe the update rules for NMF/$\mathcal{T}$/NB$_{\alpha}$, we derive new updates for NMF/$\mathcal{C}$/NB$_{\alpha}$ and we outline the relationship between the Negative Binomial and the Tweedie distribution with its special cases. 

Let $X$ be Negative Binomial distributed with mean $\mu$ and dispersion parameter $\alpha$, then  $$X \sim \text{NB}(\alpha, \frac{\mu}{\alpha + \mu})$$ and we have that $$\mathbb{E}[X] = \mu \text{ and } \text{Var}X = \mu \left(1+\frac{\mu}{\alpha} \right).$$

As for the Tweedie distribution, the Negative Binomial distribution is of great interest for various applications. In particular, when modelling count data with overdispersion it is a more appropriate model than the Poisson distribution. Examples of the applicability of the Negative Binomial distribution can be found in modelling mutational counts in cancer patients \citep{Pelizzola2023}, infectious disease outbreaks \citep{LloydSmith2005} and in quality control problems in industrial settings \citep{Albers2010}.

For a non-negative matrix $V \in \mathbb{R}_+^{M \times N}$, the model assumed by NMF is given by $V_{nm} \sim \text{NB}(\alpha, \frac{\mu_{nm}}{ \alpha + \mu_{nm}}) $ with $ \mu_{nm} = (WH)_{nm} $ for traditional NMF and $ \mu_{nm} = (V^TED)^T_{nm} $ for convex NMF. Then the log-likelihood function for NMF/$\mathcal{T}$/NB$_{\alpha}$ is 
\begin{align} \label{eq:fulllik}
    \ell(W,H;V) & = \sum_{n=1}^N \sum_{m=1}^M \Bigg\{ \log { \binom{\alpha + V_{nm}  - 1}{\alpha}} + V_{nm} \log \left( { \frac{(WH)_{nm}}{\alpha + (WH)_{nm}}}\right)  \\ \nonumber
    &  + \alpha \log \left( { 1 - \frac{(WH)_{nm}}{\alpha + (WH)_{nm}} } \right) \Bigg\},
\end{align} 
where $\binom{\alpha + V_{nm}  - 1}{\alpha}$ is the binomial coefficient which can be also expressed as $\frac{\Gamma(\alpha + V_{nm})}{\Gamma(\alpha +1)\Gamma(V_{nm})}$ for non-integer values of $\alpha$. The negative of the log-likelihood function is proportional to the following divergence:
\begin{align}
    d_N(V||WH) & = \sum_{n=1}^N \left \{  \sum_{m=1}^M V_{nm} \log \left(\frac{V_{nm}}{ (WH)_{nm}}\right) - (\alpha + V_{nm}) \log \left(\frac{\alpha + V_{nm}}{\alpha + (WH)_{nm}} \right) \right \}
    \label{eq:divmethods}
\end{align} 
assuming fixed $\alpha$. The term $\log { \binom{\alpha + V_{nm}  - 1}{\alpha}}$ in equation \eqref{eq:fulllik} is a constant that can be omitted and the constant terms $V_{nm} \log V_{nm}$, $\alpha  \log \alpha$, and $(\alpha + V_{nm}) \log (\alpha + V_{nm})$ can be added to obtain the divergence in equation \eqref{eq:divmethods}.
Following the steps in \cite{Gouvert2020} and \cite{Pelizzola2023} updates for $W$ and $H$ under this model can be derived using a MM algorithm and they are as follows:
\begin{align} \label{eq:updateWH_negbin}
   H^{t+1} = H^t \cdot \frac{\frac{W^TV}{(WH^t)}}{\frac{W^T(V + \alpha)}{(WH^t) + \alpha}}
 \text{ and }
    W^{t+1} = W^t \cdot \frac{ \frac{VH^T}{(W^tH)}}{\frac{(V + \alpha)H^T}{(W^tH) + \alpha}}.  
\end{align}

Update rules for NMF/$\mathcal{C}$/NB$_{\alpha}$ are not available in the literature, but can be derived with a similar procedure. In this case the divergence to be minimised is:
\begin{align}
    d_N(V^T||V^TED) & = \sum_{n=1}^N \left \{  \sum_{m=1}^M V^T_{mn} \log \left(\frac{V^T_{mn}}{ (V^TED)_{mn}}\right) - (\alpha + V^T_{mn}) \log \left(\frac{\alpha + V^T_{mn}}{\alpha + (V^TED)_{mn}} \right) \right \}. \label{eq:divmethods2}
\end{align} 
Since minimising the divergence in equation \eqref{eq:divmethods2} is not convex in both $E$ and $D$ simultaneously, we fix one of them at a time and only optimise the other. This results in a convex problem and an MM algorithm for minimising the divergence leads to the following updates for $E$ and $D$:
\begin{align} \label{eq:updateED_Cnegbin}
    E^{(t+1)} = E^{t}\cdot\frac{\left( V \left( \frac{V^T}{V^TE^tD}\right)D^T\right)}{\left( V \left( \frac{\alpha + V^T}{\alpha +V^TE^tD}\right)D^T\right)}
 \text{ and }
    D^{(t+1)} = D^{t}\cdot\frac{\left((V^TE)^T \left( \frac{V^T}{V^TED^t}\right)\right)}{\left( (V^TE)^T \left( \frac{\alpha + V^T}{\alpha +V^TED^t}\right)\right)}.
\end{align}
We provide the full derivation in Appendix \ref{appB:negbincinvex}. In our application, we find maximum likelihood estimates (MLEs) of $\alpha$ based on the Negative Binomial likelihood using Newton-Raphson together with the estimate of $V^TED$ from NMF/$\mathcal{C}$/Po as proposed in \cite{Pelizzola2023} for traditional NMF. This estimation procedure is described in Algorithm \ref{alg:nbCnmf_alpha} below.

\begin{algorithm}[h!]
\caption{NMF/$\mathcal{C}$/NB$_\alpha$/$K$: Estimation of $E$, $D$ and $ \alpha $} \label{alg:nbCnmf_alpha}
\begin{algorithmic}[1]
\Require{$V, K, \epsilon$}
\Ensure{$E$, $D$, $\alpha,$ } 
\State $E^{\text{NMF/} \mathcal{C} \text{/Po/$K$}}, D^{\text{NMF/} \mathcal{C} \text{/Po/$K$}} \gets$ apply NMF/$\mathcal{C}$/Po to $V$ 
\State $\alpha \gets$ Negative Binomial MLE using $E^{\text{NMF/} \mathcal{C} \text{/Po/$K$}}, D^{\text{NMF/} \mathcal{C} \text{/Po/$K$}}$ and $V$
\State Initialise $E^1,D^1$ from a random uniform distribution
\For{$t = 1, 2, \dots$} 
\State $E^{t+1} \gets E^{t}.\frac{\left( V \left( \frac{V^T}{V^TE^tD}\right)D^T\right)}{\left( V \left( \frac{\alpha + V^T}{\alpha +V^TE^tD}\right)D^T\right)}$
\linespread{2.7}\selectfont
\State $ D^{(t+1)} \gets D^{t}.\frac{\left((V^TE)^T \left( \frac{V^T}{V^TED^t}\right)\right)}{\left( (V^TE)^T \left( \frac{\alpha + V^T}{\alpha +V^TED^t}\right)\right)}$    
\linespread{2.7}\selectfont
\If{$|d_N(V||VE^{t+1}D^{t+1}) - d_N(V||VE^{t}D^{t}) | < \epsilon$}
\State \Return $ E, D \gets E^{t+1}, D^{t+1}$
\linespread{1}\selectfont
\EndIf
\EndFor
\end{algorithmic}
\end{algorithm}

\subsubsection{Relation between Negative Binomial and Tweedie NMF methods} \label{sec:tweedienegbin}
The relationship between Tweedie NMF and Negative Binomial NMF is governed by their respective variance–mean laws. The Tweedie and the Negative Binomial distribution have distinct mean-variance relations, that are controlled by their parameters. For a Tweedie distributed random variable $X$ with mean $\mu$, power $p$ and variance parameter $\sigma^2$, the mean-variance relationship is $$\text{Var} X = \sigma^2 \mu^p.$$ For a Negative Binomially distributed random variable $X$ with mean $\mu$ and variance parameter $\alpha$, the relationship is given by $$\text{Var} X = \mu + \frac{\mu^2}{\alpha}.$$  Both families have the Poisson distribution as a special case in which the variance equals the mean. This is obtained in the Tweedie distribution by setting $(\sigma^2, \mu) = (1,1)$ and in the Negative Binomial distribution by taking the limit $\alpha \rightarrow \infty$. Thus, Poisson NMF can be viewed as a limiting case of Negative Binomial NMF, as shown in \cite{Gouvert2020}, and Negative Binomial NMF naturally extends Poisson NMF by allowing for overdispersion. Tweedie NMF provides a broader generalization by parameterizing variance growth via the power parameter $p$, which enables modelling a wider range of mean–variance laws, with the Poisson being an important special case. The same rationale applies to the relationship between NMF/$\mathcal{C}$/TW$_p$ and NMF/$\mathcal{C}$/NB$_\alpha$.

\subsection{Computational aspects} 
\label{sec:computational}
This section provides details on the computational aspects of the algorithms outlined in Sections \ref{sec:tweedie} and \ref{sec:negbin}. For this analysis we used the data set of mutational counts from liver cancer patients from the Pan-Cancer Analysis of Whole Genomes (PCAWG) database \citep{Campbell2020} with $N = 260$ patients. We considered different subsets of patients to test the run time as a function of the number of observations $N$ and two alternative definitions of mutation types to show the dependency of the run time on the number of classes $M$. We show results for short and long contextual pattern around the location where the mutation is observed which is also relevant for this application domain as the context where the mutation occurs has been shown to impact mutation rates in the genome \cite{Lindberg2019}.

\begin{figure}[h!]
    \centering
    \includegraphics[width=\linewidth]{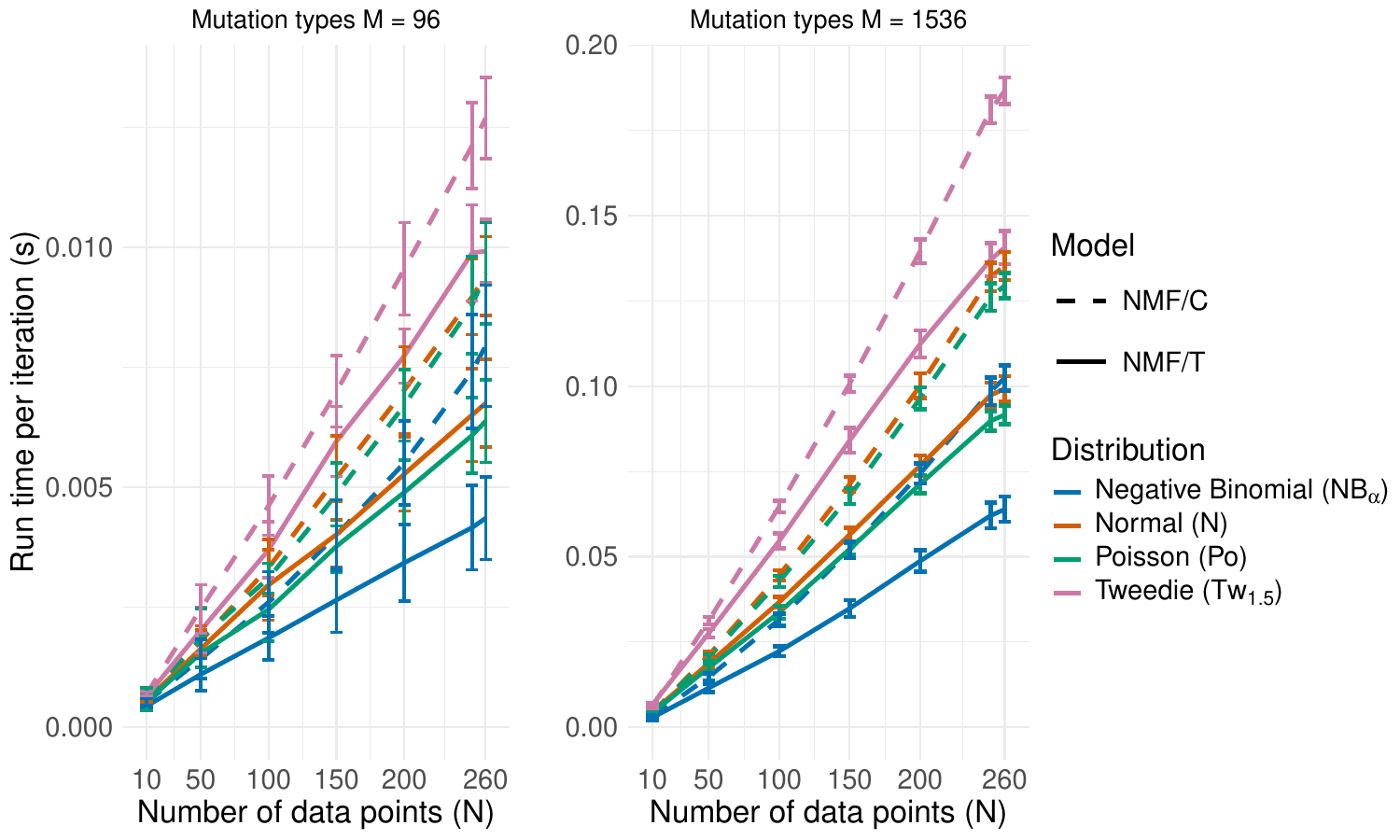}
    \caption{Run time analysis per iteration. The run time in seconds is plotted against the number of data points $N$ for the different methods and distributional assumptions and number of dimensions M. Results are averaged over 100 iterations. The rank of the factorisation is K = 5 as estimated in Section \ref{sec:liver}, and the values of the dispersion parameter $\alpha$ are as given in \ref{fig:residualplotliver}, being 45.21 and 36.44 for traditional and convex NMF respectively.}
    \label{fig:runtime}
\end{figure}

For traditional NMF the updates can be found in Table \ref{tab:normalpoissonupdates} and in equations \eqref{eq:NMF-TW-H-W} and \eqref{eq:updateWH_negbin}. Each of these individual update steps of $W$ or $H$  has a computational complexity of $\mathcal{O}(MNK)$. As can be seen from their structure, each update amounts to a number of matrix multiplications which are the main contributors to this computational cost.  The convex NMF updates are described in Table \ref{tab:normalpoissonupdates} and equations \eqref{eq:CNMF-TW-E-D} and \eqref{eq:updateED_Cnegbin}. As before, the computational cost is greatly affected by all the matrix multiplications needed to compute each of these updates, however due to the structure of convex NMF, an update step of $E$ or $D$ has a computational complexity of $\mathcal{O}(MN^2K)$. 
This is consistent across all considered distributions, as they all share a common MM-based update structure. We provide results on the run time in Figure \ref{fig:runtime}. 

As can be seen in both panels, the runtime of traditional NMF depends linearly on $N$ and NMF/$\mathcal{C}$ is usually slightly slower than NMF/$\mathcal{T}$. Furthermore, comparing the two panels of Figure \ref{fig:runtime} the linear dependency in $M$ can also be noticed with a 10 fold difference in the number of mutation types corresponding to a 10 fold difference in runtime.
NMF/$\mathcal{T}$/NB$_\alpha$ is the fastest of all methods. We also observe that NMF/$\mathcal{T}$/TW$_p$ and NMF/$\mathcal{C}$/TW$_p$ are slower for certain values of $p$, as the runtime of NMF/$\mathcal{T}$/TW$_{1.5}$ and NMF/$\mathcal{C}$/TW$_{1.5}$ is higher than their Normal and Poisson counterparts. This is due to the computational costs of raising matrices to the power $p$ for each iteration: when $p \in \{0,1\}$, corresponding to the Normal and Poisson models, no calculations are performed (see updates in Table \ref{tab:normalpoissonupdates}), whereas for the general Tweedie distribution the updates involve raising matrices to the power $p$ as shown in equations \eqref{eq:NMF-TW-H-W} and \eqref{eq:CNMF-TW-E-D} which slows down each iteration. 

Furthermore, NMF/$\mathcal{T}$/TW$_p$ and  NMF/$\mathcal{C}$/TW$_p$ require additional computational time to estimate $p$, and NMF/$\mathcal{T}$/NB$_\alpha$ and  NMF/$\mathcal{C}$/NB$_\alpha$ also require additional time to estimate $\alpha$. As an example, we report computational times (in seconds) averaged over 500 iterations to estimate $\alpha$ and $p$ for the liver mutational counts data. On a standard laptop, the $\alpha$ estimation is very fast, taking just $\sim45$ seconds on average, however estimating $p$ takes around 3 minutes. 

\section{Results} \label{sec:results}
%scale invariant symmetric relative change for stopping
%
We conduct a residual analysis to compare the goodness of fit of four distributions for each of NMF/$\mathcal{T}$ and NMF/$\mathcal{C}$. Residuals provide a natural diagnostic for evaluating whether the assumed noise model is compatible with the empirical variability in the data. Specifically, we examine plots of raw residuals against fitted values to compare Gaussian, Poisson, Negative Binomial, and Tweedie models. Residuals are defined as $R_{nm} = V_{nm} - \hat{V}_{nm}$ that is, the deviation of the observed data from the fitted values implied by the model. To evaluate the adequacy of each distributional assumption, we plot these residuals against the expected means $\hat{V}_{nm}$ = ($\hat{W}\hat{H})_{nm}$.

\subsection{Mutational count data from liver cancer patients} \label{sec:liver}

\begin{figure}[h!]
    \centering
    \includegraphics[width=0.8\linewidth]{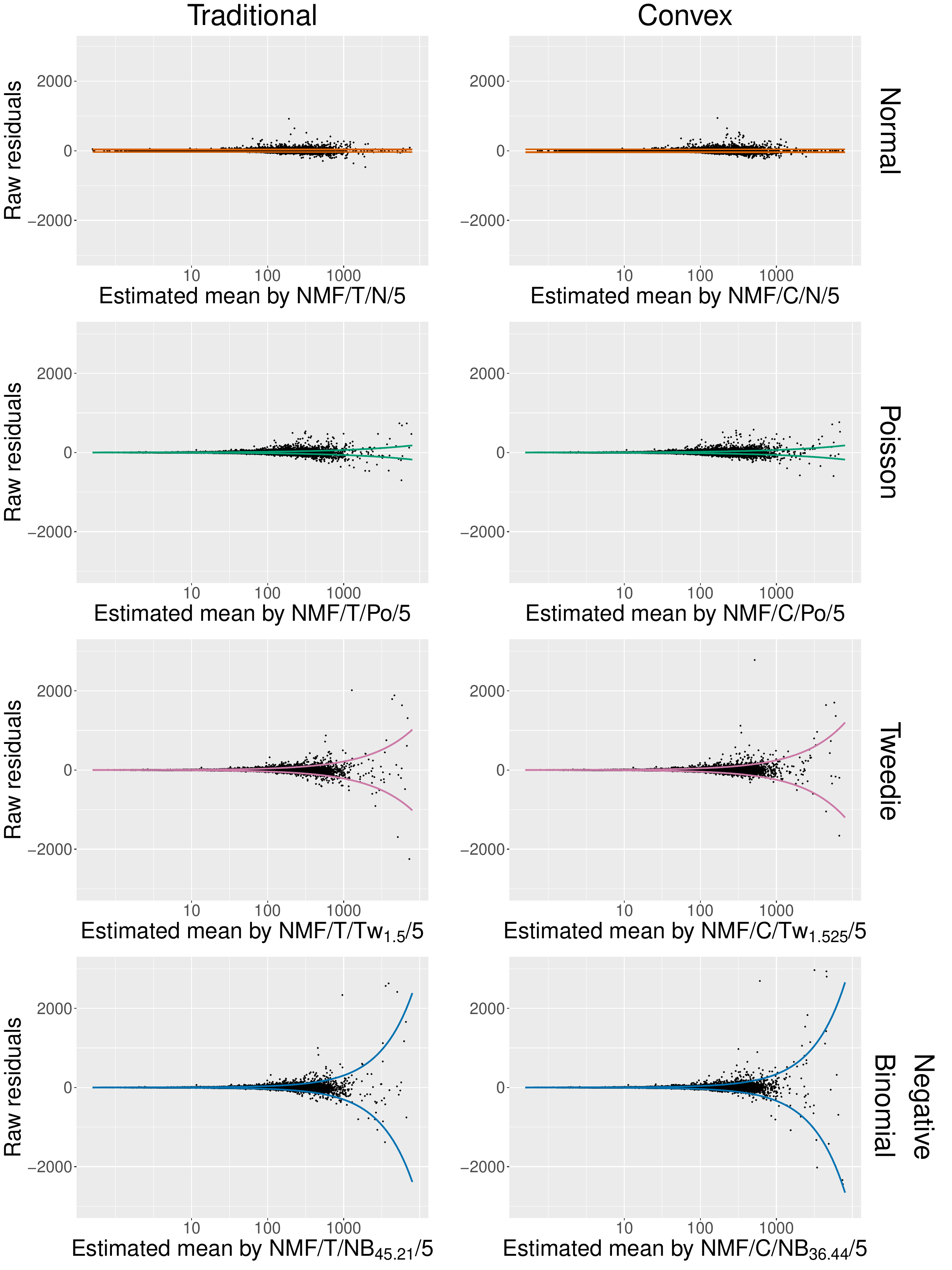}
    \caption{Residual analysis of NMF/$\mathcal{T}$ (on the left) and NMF/$\mathcal{C}$ (on the right) under the Normal, Poisson, Tweedie, and Negative Binomial distributions applied to the mutational count data of liver cancer patients. In each panel, the coloured lines are 2 standard deviations of the assumed model.}
    \label{fig:residualplotliver}
\end{figure}

In Figure \ref{fig:residualplotliver}, the fit of various models is assessed for a dataset of mutational counts in liver cancer.

\begin{figure}[h!]
    \centering
    \includegraphics[width=0.9\linewidth]{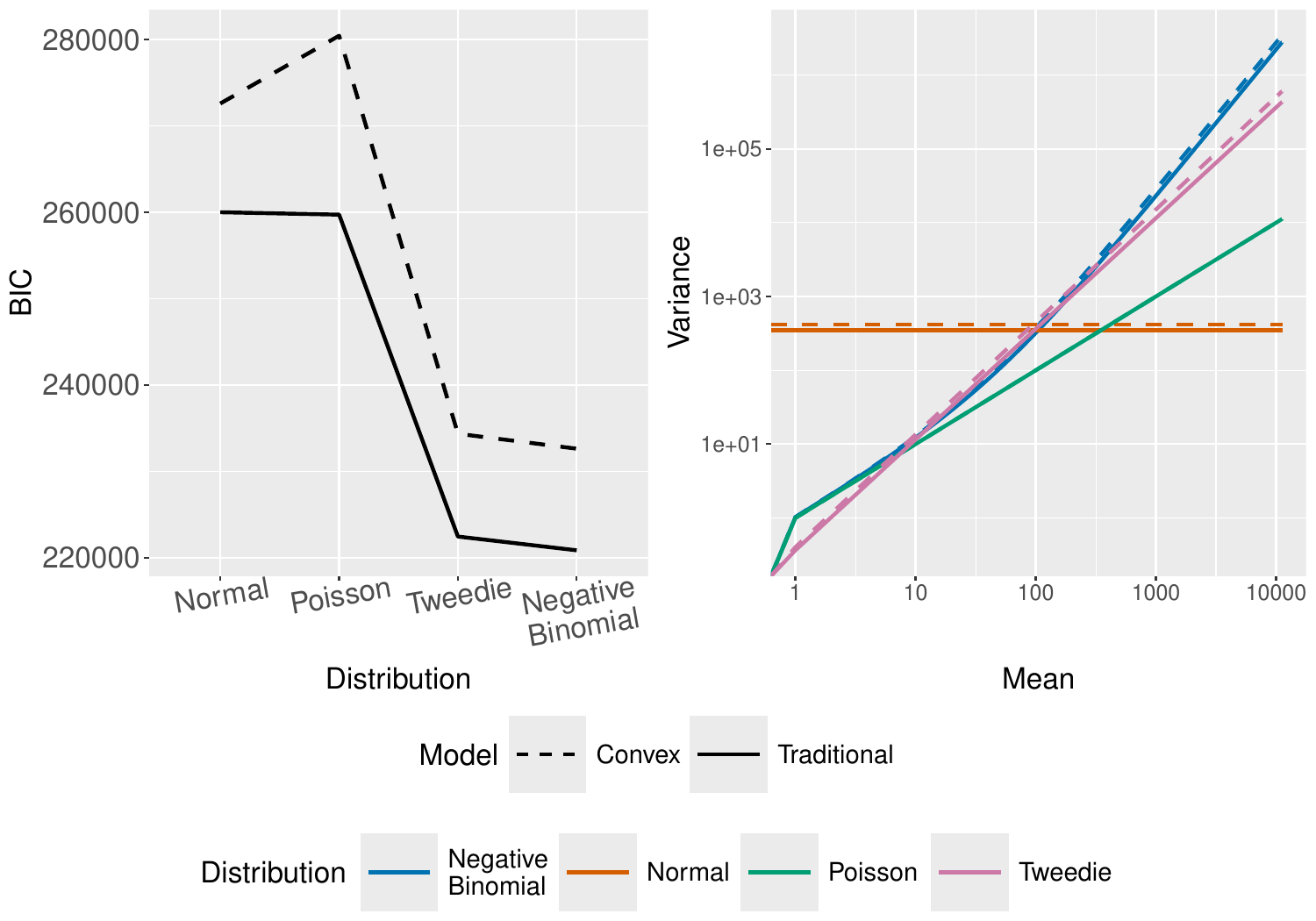}
    \caption{Comparison of BIC values and mean-variance relationships across the 8 NMF models, applied to the mutational count data of liver cancer patients.}
    \label{fig:meanvarBICliver}
\end{figure}

This is a subset of the available data from the Pan-Cancer Analysis of Whole Genomes (PCAWG) database \citep{Campbell2020}, thus it corresponds to one of the largest available data sets for a single cancer type. For the $N = 260$ patients, the counts of each of the $M = 96$ mutation types are recorded. The same rank ($K = 5$) is used for all methods, and it is estimated  by cross validation using the $\texttt{SigMoS}$ package \citep{Pelizzola2023}. 
The number of free parameters in the traditional NMF models is $NK + M(K-1)$ plus 1 for NMF/$\mathcal{T}$/TW$_p$ and NMF/$\mathcal{T}$/NB$_\alpha$ corresponding to the estimation of $p$ and $\alpha$ respectively. This amounts to 1684 parameters for the Normal and Poisson NMF and 1685 parameters for the Tweedie and Negative Binomial NMF.  Following the same rationale, for convex NMF we need to estimate $NK + N(K-1) = 2340$ (with Normal and Poisson cost functions) or $NK + N(K-1) + 1 = 2341$ free parameters (with Tweedie and Negative Binomial cost functions) which is much higher than those for traditional NMF models.

The plot of BIC values in Figure \ref{fig:meanvarBICliver} (left) shows that traditional NMF has lower BIC values than convex NMF across all distributional assumptions. For both methods of NMF, we see that the Tweedie and Negative Binomial models outperform the Normal and Poisson models, with NMF/$\mathcal{T}$/NB$_{45.21}$/5 and NMF/$\mathcal{C}$/NB$_{36.44}$/5 providing the best fit for traditional and convex NMF respectively. 
% Mean-var plot shows that this is because they model similar meanvar relation
In Figure \ref{fig:meanvarBICliver} (right) we show the estimated mean-variance relationship under each of these 8 models. For a given distributional assumption, the mean-variance relationships of traditional and convex NMF are very similar with comparable estimates of both the power parameter $p$ in Tweedie NMF and the dispersion parameter $\alpha$ for Negative Binomial NMF. The mean-variance relationship of Tweedie and Negative Binomial models are similar, and both grow faster than the mean. In the case of NMF/$\mathcal{T}$/Po and NMF/$\mathcal{C}$/Po the mean-variance relationships of the traditional and convex NMF models are identical. 

Residual diagnostics in Figure \ref{fig:residualplotliver} confirm that the distributional assumptions from Normal and Poisson NMF respectively provide a poor fit, , reflecting their inability to accommodate overdispersion. In contrast, the Tweedie and Negative Binomial models provide similar and substantially improved residual behaviour; this underlines the importance of making more flexible models available for data analysis, such as the Tweedie and Negative Binomial models we are proposing. The better fit provided by the more flexible models is in agreement with previous results on mutational counts data from cancer patients showing that the Negative Binomial is a more suitable model assumption than the commonly chosen Poisson distribution \citep{Lyu2020, Pelizzola2023} as in these data the variance is usually greater than the mean. 

\begin{figure}[h!]
    \centering
     \includegraphics[width = \textwidth]{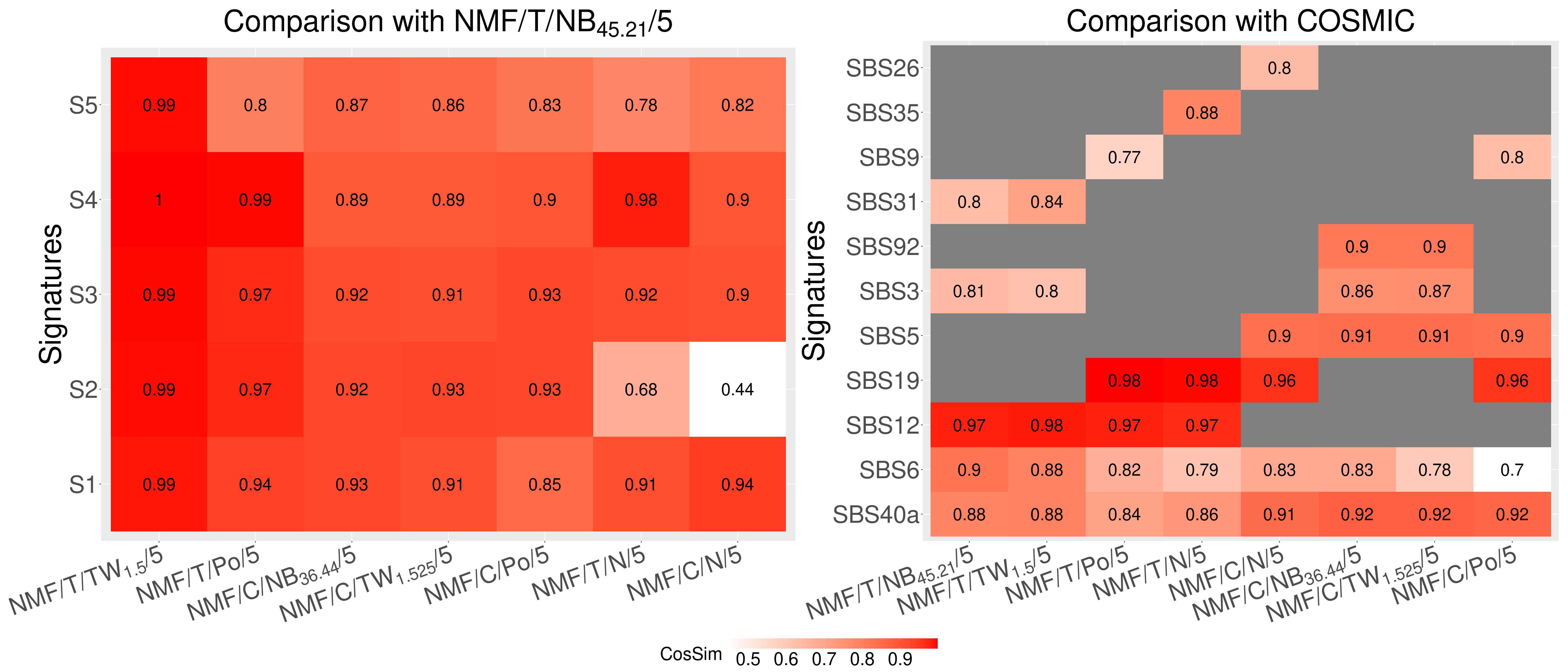}
    \caption{Quality of estimated signatures for mutational count data of liver cancer patients. Each method is compared to NMF/$\mathcal{T}$/NB$_{45.21}$/5 (left) and to the COSMIC signatures (right). We show the quality of the estimated signatures measured by cosine similarity for all methods.}
    \label{fig:signaturecomparisonliver}
\end{figure}

We also checked the quality of the estimated signatures for all methods. We calculated the cosine similarity between the signatures estimated by NMF/$\mathcal{T}$/NB$_{45.21}$/5 as this is the method with the smallest BIC value and all other methods and then compared all methods to the publicly available COSMIC signatures \citep{Tate2019} which are curated signatures derived from large cohorts of cancer patients. 
Figure \ref{fig:signaturecomparisonliver} shows that traditional NMF with the Tweedie and Poisson distribution and NMF/$\mathcal{C}$/NB$_{36.44}$/5 estimated signatures which are quite similar to those estimated by NMF/$\mathcal{T}$/NB$_{45.21}$/5. In particular, NMF/$\mathcal{T}$/NB$_{45.21}$/5 and NMF/$\mathcal{T}$/TW$_{1.5}$/5 have an average cosine similarity of $0.992$.
The comparison with COSMIC shows that all methods estimate signatures that have been previously found in liver cancer patients \citep{Alexandrov2020} with NMF/$\mathcal{T}$ with Negative Binomial and Tweedie cost functions and NMF/$\mathcal{C}$ with Negative Binomial and Normal cost functions having a cosine similarity higher than 0.8 for all estimated signatures. We note that a  cosine similarity of 0.8 has been used as threshold in \cite{Pei2020} to group signatures that were considered equivalent to each other, however this measure tends to favour signatures with high peaks \citep{sidorov2014soft}. Most models show high cosine similarity to at least four of the matched COSMIC signatures and these results confirm that the Negative Binomial distribution is a robust model here being the only model with cosine similarity higher than 0.8 with both traditional and convex NMF. Furthermore, traditional NMF methods are able to recover SBS12 almost perfectly and this is the main signature associated to liver cancer in the COSMIC database. More details on the aetiologies of the different signatures identified here can be found in \citep{Alexandrov2020}. Matching between the COSMIC database and the estimated signatures is done using the Hungarian algorithm.

\subsection{Topic modelling of newsgroups data} \label{sec:newsgroups}
\begin{figure}[h!]
    \centering
    \includegraphics[width=0.8\linewidth]{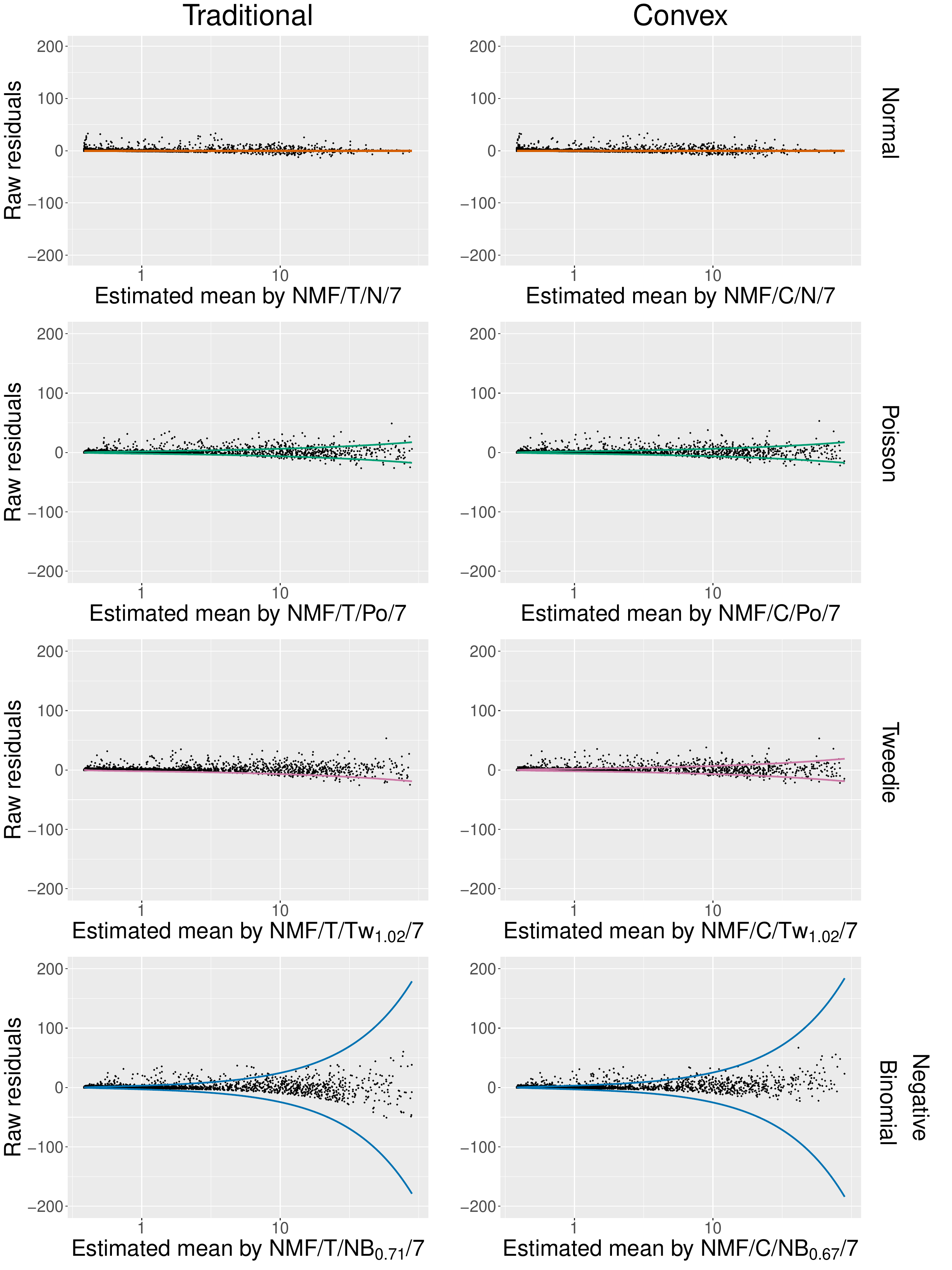}
    \caption{Residual analysis of NMF/$\mathcal{T}$ (on the left) and NMF/$\mathcal{C}$ (on the right) under the Normal, Poisson, Tweedie, and Negative Binomial distributions applied to the word count data of 500 documents. In each panel, the coloured lines denote twice the standard deviation of the assumed model. The residual plots are downsampled from 3177000 to 100000 points.}
    \label{fig:residualplotnewspaper}
\end{figure}

In Figure \ref{fig:residualplotnewspaper}, a similar residual analysis is conducted on text data from newsgroups documents from \cite{Lang1995}. The documents are readily available and can be found at \url{http://qwone.com/~jason/20Newsgroups/} (last accessed 02.09.2025). We restrict our analysis to 500 documents regarding sport, religion and politics. In particular, we consider a random subset of documents from three different subtopics ("middle east", "hockey" and "Christianity"), we removed connecting words (such as "and", "or") and only consider words that are present in at least 70 documents. 

\begin{figure}[h!]
    \centering
    \includegraphics[width=\linewidth]{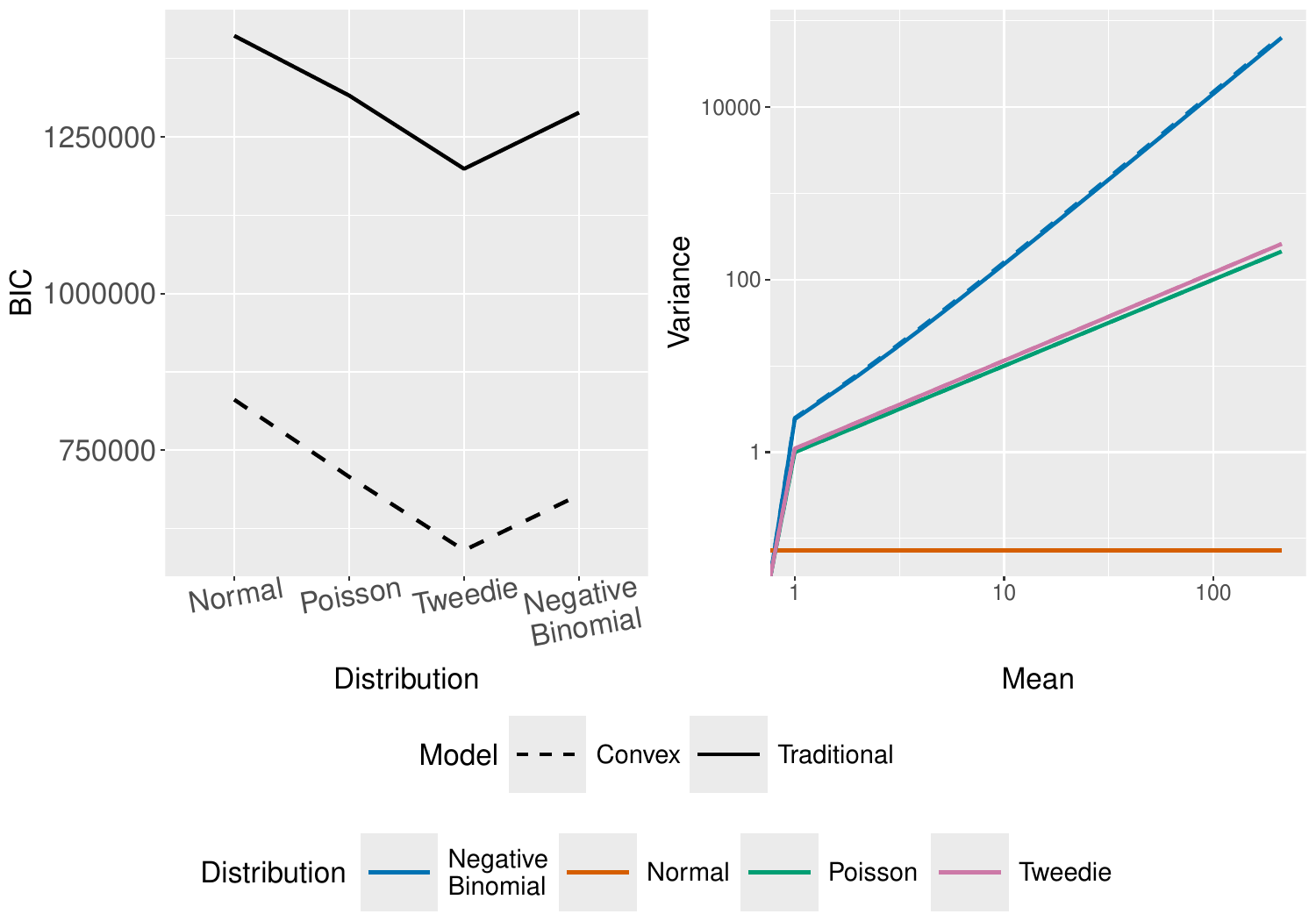}
    \caption{Comparison of BIC values and mean-variance relationships across the 8 NMF models on the word count data of 500 documents.}
    \label{fig:meanvarBICplotnewspaper}
\end{figure}

% Residuals newspaper data

This results in a data set recording counts of the appearance of 6354 words in 500 documents. The same rank ($K = 7$) is used for all methods and it is chosen following the results in \citep{Carbonetto2025}. The number of free parameters in the traditional NMF models is 41624 ($+ 1$ for the estimation of $\alpha$ and $p$ in the Negative Binomial and Tweedie models, respectively), which is more than 6 times higher than the 6500 ($+ 1$) parameters estimated in the convex NMF models.

Figure \ref{fig:residualplotnewspaper} shows the residual analysis for all fitted models and reports the corresponding BIC values. The model with lowest BIC is NMF/$\mathcal{C}$/TW$_{1.02}$/7. The second lowest BIC is obtained by NMF/$\mathcal{C}$/NB$_{0.45}$/7, followed closely by NMF/$\mathcal{C}$/Po/7 and both provide a reasonably good fit to the data. On the contrary, due to the high level of sparsity of this data set, the Normal distribution is not suitable in this scenario as shown in the top panel of Figure \ref{fig:residualplotnewspaper}. 

For this application, the fitted Tweedie and Poisson models yield nearly indistinguishable mean-variance relationships as shown in Figure \ref{fig:residualplotnewspaper} (bottom-right panel). This is explained by the estimated Tweedie power parameter $p = 1.02$, which places the model very close to the Poisson regime (corresponding to the Tweedie distribution with $p = 1$ as described in Section \ref{sec:normalnmf}). In contrast, the Negative Binomial model corresponds to a different mean–variance relationship, with the estimated dispersion parameter $\alpha = 0.45$ inducing substantially higher variance than the Poisson model. We note that varying $\alpha$ over a range of values that would bring the Negative Binomial mean-variance relationship close to the one from the Poisson distribution produces very similar results in terms of factorisation, but decreases the overall goodness of fit. The likelihood profile for $\alpha$ in Figure \ref{fig:alphaprofile} shows that the dispersion parameter may admit a wide confidence interval in this setting. 
Comparing traditional and convex NMF further highlights the interaction between model structure and data sparsity. In this application, convex NMF achieves comparable fit with substantially fewer parameters, suggesting that the convexity constraint acts as an effective form of regularization in high-dimensional sparse settings and reduces overfitting without losing explanatory power in the extracted features. 
\begin{figure}[h!]
    \centering
    \includegraphics[width=\linewidth]{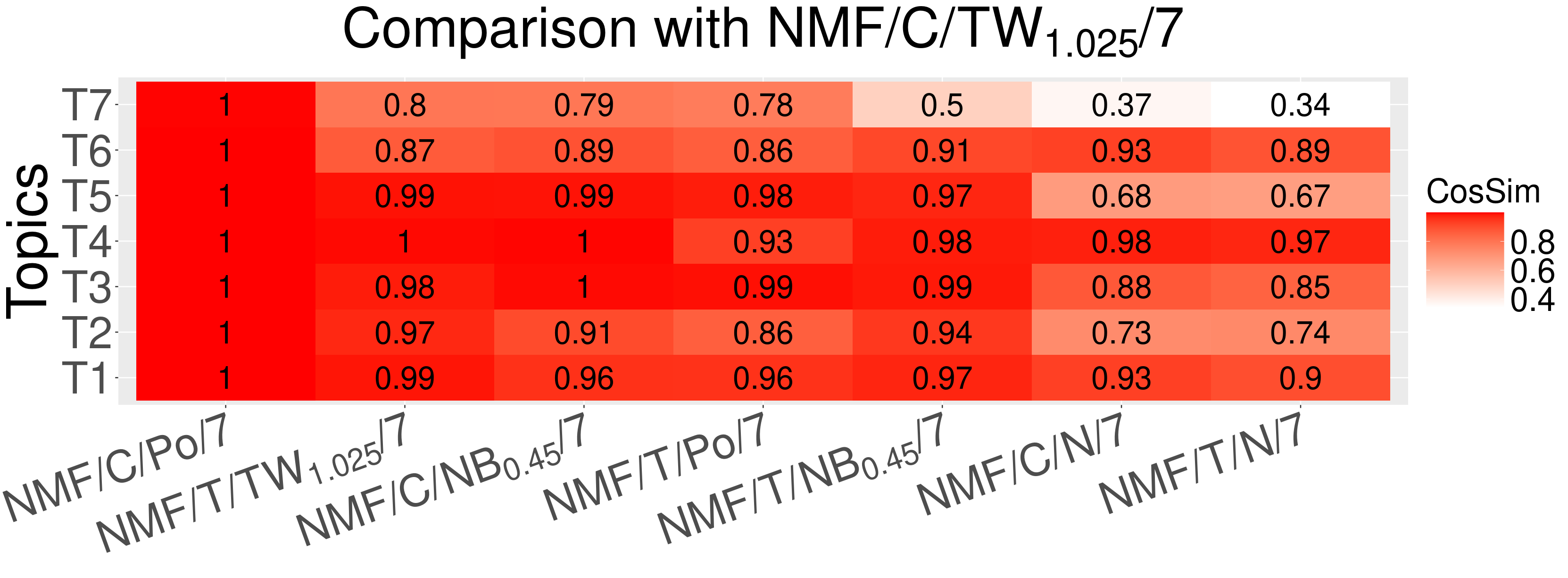}
    \caption{Quality of estimated features for text count data of newsgroups posts. Each method is compared to NMF/$\mathcal{C}$/TW$_{1.026}$/7. We show the quality of the estimated features measured by cosine similarity for all methods.}
    \label{fig:signaturecomparisonnewsgroups}
\end{figure}

Cosine similarities of the estimated features compared to the best model (NMF/$\mathcal{C}$/TW$_{1.02}$/7) are shown in Figure \ref{fig:signaturecomparisonnewsgroups}. These results confirm the superiority of convex NMF in this setting: the features obtained from the best-performing model are highly concordant with those obtained from convex NMF with the Poisson and Negative Binomial models, whereas the traditional NMF features extracted with Poisson and Negative Binomial are slightly less concordant. The cosine similarities of the features extracted with the Normal model are much lower than the rest of the observed cosine similarities with both traditional and convex NMF showing once again that this model is not a good choice here.  Moreover, we also observed that the estimated features can be clearly associated with specific topical domains, and the corresponding document weights align well with the known document labels. Topical domains estimated by NMF/$\mathcal{C}$/TW$_{1.02}$/7 are shown in Figure \ref{fig:convextweedietopics} in Appendix \ref{app:D}. Figure \ref{fig:convextweedieweights} in Appendix \ref{app:D} shows the corresponding document weights where the alignment between the known documents labels and the estimated topics is highlighted by ordering the documents with respect to their original labels. We note that cosine similarities here are very high, this is mainly due to the definition of this measure favouring high peaks in the features which are more pronounced for sparse data.

\section{Discussion} \label{sec:discussion}
NMF is a widely used dimensionality reduction technique for extracting latent features from large data set. It relies on a chosen distributional assumption which greatly affects the shape of the extracted features and the model fit. In this paper, we propose a unified framework for both traditional and convex NMF with various underlying cost functions corresponding to different mean–variance regimes and introduce new multiplicative updates for convex NMF with Poisson, Tweedie and Negative Binomial cost functions. We show how classical Gaussian and Poisson models emerge as special cases of the more general Tweedie framework which provide greater flexibility for overdispersed data together with the Negative Binomial formulation. 

We provide an efficient implementation of traditional and convex NMF with all considered cost functions based on multiplicative updates derived with MM-algorithms to ease the applicability of these methods to large data sets. This allows to easily perform new NMF analysis with the suitable cost function for any given data set and to compare the different distributional assumptions on new data by repeating the residual analysis from this manuscript on the data of interest.

We consider a data set of mutational counts data from liver cancer patients (Section \ref{sec:liver}) and one of word count data from newsgroups posts (Section \ref{sec:newsgroups}) and apply all models to both data sets. We show that the choice of noise model for both traditional and convex NMF has a substantial impact on model fit and should be guided by the empirical mean–variance relationship of the data. Gaussian-based NMF is most suitable when variability is approximately constant, while Poisson or Negative Binomial models are preferable for count data and in particular the Negative Binomial model becomes essential when overdispersion is present and the variance increases with the mean. Tweedie models offer additional flexibility, but may come at the cost of slower convergence, reflecting the more complex minimisation problem. These observations emphasize the importance of explicitly modelling mean–variance structure in matrix factorisation problems and suggest that treating NMF as a statistical model rather than an algorithmic procedure provides valuable statistical insight. 

Within a fixed distributional family, we observe that on the liver count data set, traditional NMF achieves systematically lower BIC values when compared to the convex NMF factorisation results, whereas we observe the opposite behaviour on the text count data set. A closer inspection of the convex NMF results shows that this model achieves likelihood values comparable to those of traditional NMF while using substantially fewer parameters. This suggests that convex-NMF leads to improved BIC in settings characterized by high dimensionality and sparsity and indicates that convexity can act as an effective form of regularization in high-dimensional, sparse settings.

This work underscores the importance of aligning NMF model choices with the statistical properties of the data. Based on the results presented in this paper, we suggest to first examine empirical mean–variance relationships and sparsity patterns, as these provide essential guidance for selecting an appropriate model. Further work is needed to better understand the theoretical properties of convex NMF under different noise models, particularly in terms of identifiability and uniqueness of the solutions as previously investigated for traditional NMF \citep{Laursen2022}. Developing data-driven procedures that jointly select the variance function and structural constraints, allowing the model to adjust automatically to changes in the data would also be relevant in applied contexts. Such developments would further strengthen the role of NMF as a statistically principled modelling tool.

We have implemented all models considered in this manuscript within the same R package nmfgenr which is available at \url{https://github.com/MartaPelizzola/nmfgenr}.

\section*{Acknowledgement}
MP and AH acknowledges funding of the Novo Nordisk Foundation (Grant number NNF21OC0069105). ESJ and AH acknowledges funding of the Novo Nordisk Foundation (Grant number NNF22OC0079957). Some of the computing for this project was performed on the GenomeDK cluster. We would like to thank GenomeDK and Aarhus University for providing computational resources and support that contributed to these research results.

\newpage
\begin{appendices}

\section{Power fitting for Tweedie NMF} \label{app:A}
\begin{figure}[h!]
    \centering
    \includegraphics[width=\linewidth]{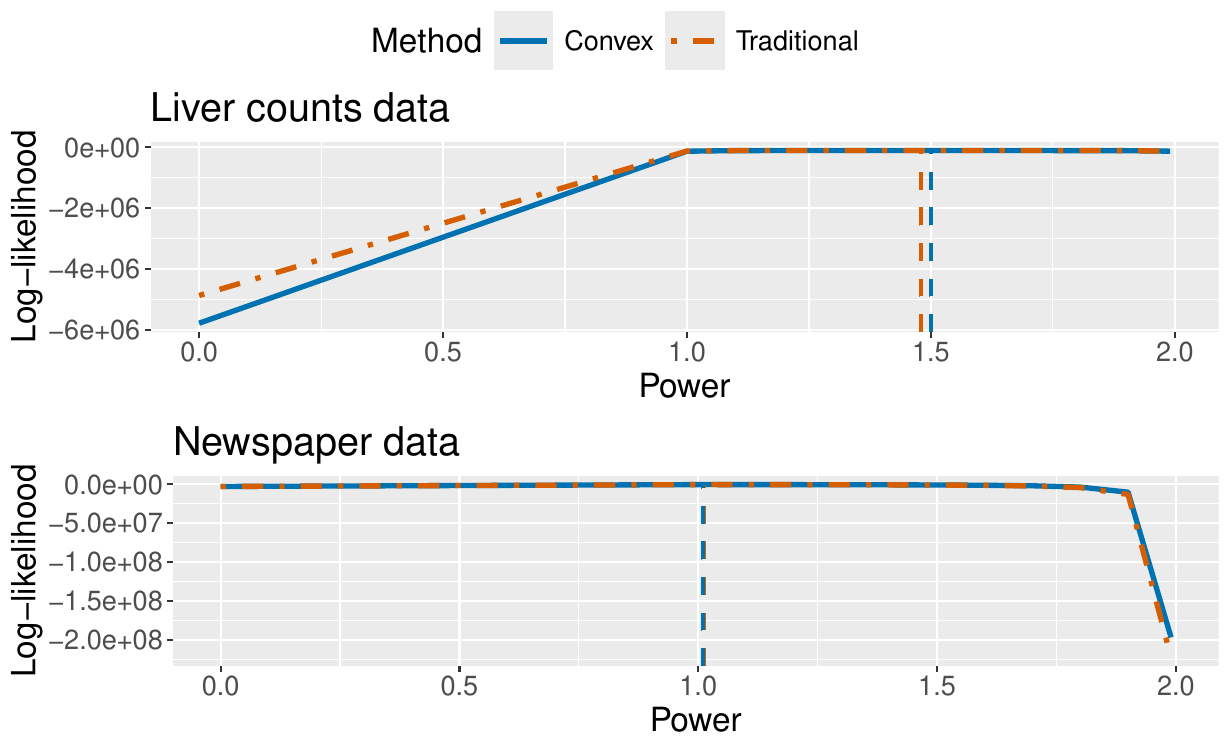}
    \caption{Fitting of power parameter for datasets of newsgroups word counts and mutational counts from liver cancer patients.}
    \label{fig:profilepower}
\end{figure}
The power and dispersion parameters for Tweedie NMF are fitted by the profile log-likelihood function. For each value of $p$, NMF is run using the corresponding model, and the dispersion parameter is fitted to the data, using the reconstructed data $\hat{V}$ as the mean. The log-likelihood is calculated, and the power maximising the log-likelihood is chosen. As the Tweedie distribution is undefined for $p \in (0,1)$ we consider values of the power parameter in $0\ \cup \ [1, 2]$ if the data isn't sparse, and $0\ \cup \ [1, 2)$ if there is sparsity. The dispersion parameter is generally fitted using Pearson residuals, however this is not suitable for sparse data \citep{dunn2018generalized}. Therefore for sparse data both the power and dispersion parameters are fitted by the profile log-likelihood function. 

In Figure \ref{fig:profilepower} we show the profile log-likelihood function for the liver cancer mutational counts data (upper panel) and the newsgroups word count data (lower panel).
For the liver counts dataset we notice that the profile log-likelihood function falls at $p = 1$. This is due to the dispersion parameter being fixed to $\sigma^2 = 1$ for the Poisson distribution, but being able to be fitted freely for $p = 0$ and $p > 1$. The optimum value of the power parameter is approximately $p = 1.5$ and is slightly higher for convex NMF than traditional NMF, although the curve of the profile log-likelihood function is quite flat in this region.
The optimum power for the newsgroups dataset is $p = 1.01$. We note that in Figure \ref{fig:residualplotnewspaper}, the models following the Tweedie distribution with power $p > 1$ outperform the Poisson model, however the profile log-likelihood function is also very flat in this region, and a Poisson model may also be suitable.

\section{Fitting of dispersion parameter for Negative Binomial NMF}
\begin{figure}[h!]
    \centering
    \includegraphics[width=\linewidth]{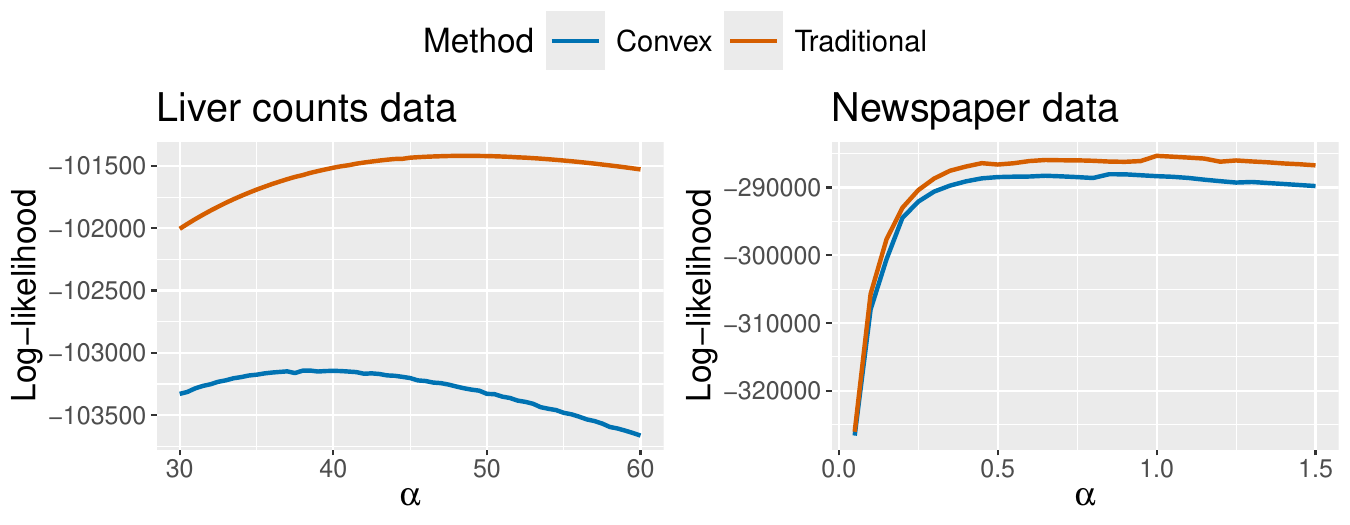}
    \caption{Profile likelihood function of $\alpha$ applied to datasets of mutational counts from liver cancer patients and newsgroups word counts.}
    \label{fig:alphaprofile}
\end{figure}

The dispersion parameters $\alpha$ for NMF/$\mathcal{T}$/NB$_\alpha$ and NMF/$\mathcal{C}$/NB$_\alpha$ are estimated via Negative Binomial maximum likelihood using the procedure outlined in Algorithm \ref{alg:nbCnmf_alpha}. Figure \ref{fig:alphaprofile} shows the profile log-likelihood function of the dispersion parameter for convex and traditional NMF on the liver cancer mutation (left) and newgroups text (right) datasets. The estimates of the dispersion parameter obtained by Algorithm \ref{alg:nbCnmf_alpha} are (36.44, 45.21) for convex and traditional NMF on the liver counts dataset, and (0.67, 0.71) for convex and traditional NMF on the newsgroup dataset. The profile likelihood curve is very flat for both of these datasets in the chosen region of $\alpha$, suggesting that a range of values of $\alpha$ could be suitable choices.

\section{Derivation of updates for Negative Binomial convex NMF} \label{appB:negbincinvex}

The Majorize-Minimisation (MM) algorithm \citep{lange2020algorithms} is used to derive multiplicative updates for NMF/$\mathcal{C}$/NB$_\alpha$. The resulting NMF method is an alternating procedure, in which  $E,D$ are initialised with random values, and updated in turn so that $E$ is updated with respect to $D$ and $D$ is updated with respect to $E$. This is iterated until the cost function converges. The columns of $E$ are normalised such that the columns of $V^TE$ sum to 1 and the rows of $D$ are multiplied by corresponding constants such that the product $V^TED$ is unchanged.
% The derivation for NB-NMF in \citep{Pelizzola2023} is similar. 

The negative log-likelihood of the negative binomial distribution is proportional to the following divergence:

\begin{align*}
    F(E,D;V) &= \sum_{n=1}^N \sum_{m=1}^M \left[ V^T_{mn}\ln \left( \frac{V^T_{mn}}{(V^TED)_{mn}} \right) - (\alpha_n + V^T_{mn}) \ln \left( \frac{\alpha_n +V^T_{mn}}{\alpha_n +(V^TED)_{mn}} \right) \right] \\
    &= \sum_{n=1}^N \sum_{m=1}^M \left[ V^T_{mn}\ln \left(V^T_{mn}\right) - V^T_{mn} \ln \left( (V^TED)_{mn}  \right) - (\alpha_n + V^T_{mn}) \ln \left( \frac{\alpha_n +V^T_{mn}}{\alpha_n +(V^TED)_{mn}} \right)  \right].
\end{align*}
Recall $\alpha \in \mathbb{R}^+_{1\times N}$ is a vector of variance parameters. Taking one value for all of $\alpha$, i.e. $\alpha_1 = \alpha_2 =\ldots = \alpha_N$ gives a dataset-specific variance parameter. Varying the values allows for patient-specific values of the variance parameter within a dataset.
\subsection{Update of E matrix}We derive an auxiliary function to majorize the divergence.
The auxiliary function $G(E,E^t)$ should fulfil the properties:
\begin{enumerate}[label=(\roman*)]
    \item $G(E,E^t) \geq F(E; D, V) \ \forall E \in \mathbb{R}^+_{N\times K}$
    \item $ G(E,E) = F(E;D,V) .$
\end{enumerate}

The first term is independent of $E$ and $D$. We majorize the second term by the log sum inequality:
\begin{align*}
    - \ln \left( (V^TED)_{mn}  \right) &= - \ln \left(\sum_{n'=1}^N \sum_{k=1}^K V^T_{mn'}E_{n'k}D_{kn} \right) \\
    & \leq - \sum_{n'=1}^N \sum_{k=1}^K c_{n'k} \ln \left( \frac{V^T_{mn'}E_{n'k}D_{kn} }{c_{n'k}} \right)
\end{align*} 

where we choose $c_{n'k} = \frac{V^T_{mn'}E^t_{n'k}D_{kn} }{(V^TE^tD)_{mn}}$ such that $\sum_{n'=1}^N \sum_{k=1}^K c_{n'k} = 1$. This applies Jensen's inequality to a double sum, over both $n'$ and $k$, but can be shown to be equivalent to summing over a single variable $j = 1\dots,n'k$ with a simple mapping.

Thus 
\begin{align*}
F(E^t; D, V) \leq \sum_{n=1}^N \sum_{m=1}^M & \left[ V^T_{mn}\ln \left(V^T_{mn}\right) - V^T_{mn} \sum_{n'=1}^N \sum_{k=1}^K c_{n'k} \ln \left( \frac{V^T_{mn'}E_{n'k}D_{kn} }{c_{n'k}} \right) \right. \\ 
& \left. - (\alpha_n + V^T_{mn}) \ln \left( \frac{\alpha_n +V^T_{mn}}{\alpha_n +(V^TED)_{mn}} \right)  \right] (\ast)
\end{align*}

Then we majorize the final term using the concavity of the log function:
\begin{align*}
    &\forall \ \ (V^TED)_{mn}, (V^TE^tD)_{mn}, \alpha_n > 0, \quad \\ & \ln(\alpha_n + V^TED_{mn})
    \leq \ln (\alpha_n +(V^TE^tD)_{mn}) + \frac{1}{\alpha_n + (V^TE^tD)_{mn}}((V^TED)_{mn} - (V^TE^tD)_{mn}).
\end{align*}

Finally we have 

\begin{align*}
    F(E^t; D, V) &\leq (\ast) \\
    \leq \sum_{n=1}^N \sum_{m=1}^M &  \left[ V^T_{mn}\ln \left(V^T_{mn}\right) - V^T_{mn} \sum_{n'=1}^N \sum_{k=1}^K c_{n'k} \ln \left( \frac{V^T_{mn'}E_{n'k}D_{kn} }{c_{n'k}} \right) \right. \\
    & \left. + (\alpha_n + V^T_{mn}) \ln \left( \frac{\alpha_n + ( V^TE^tD)_{mn}}{\alpha_n +V^T_{mn}}  \right)  + \frac{\alpha_n + V^T_{mn} }{\alpha_n + (V^TE^tD)_{mn} }(( V^TED)_{mn} - ( V^TE^tD)_{mn})\right] \\
    &= G(E, E^t).
\end{align*}
That property (i) of the auxiliary function is satisfied follows from the construction; we show here the second: $G(E,E) = F(E; D, V)$.
\begin{align*}
    G(E,E) &=  \sum_{n=1}^N \sum_{m=1}^M \left[ V^T_{mn}\ln \left(V^T_{mn}\right) - V^T_{mn} \sum_{n'=1}^N \sum_{k=1}^K \frac{V^T_{mn'}E_{n'k}D_{kn} }{(V^TED)_{mn}} \ln \left( \frac{V^T_{mn'}E_{n'k}D_{kn} }{\frac{V^T_{mn'}E_{n'k}D_{kn} }{(V^TED)_{mn}}} \right) \right. \\
    & \qquad \qquad\qquad\left. + (\alpha_n + V^T_{mn}) \ln \left( \frac{\alpha_n + (V^TED)_{mn}}{\alpha_n +V^T_{mn}}  \right)  + \frac{\alpha_n + V^T_{mn} }{\alpha_n +(V^TED)_{mn} }((V^TED)_{mn} - (V^TED)_{mn})\right] \\
    &= \sum_{n=1}^N \sum_{m=1}^M \left[ V^T_{mn}\ln \left(V^T_{mn}\right) - V^T_{mn} \sum_{n'=1}^N \sum_{k=1}^K \frac{V^T_{mn'}E_{n'k}D_{kn} }{(V^TED)_{mn}} \ln \left( (V^TED)_{mn} \right)  \right. \\
    & \qquad \qquad\qquad \left. + (\alpha_n + V^T_{mn}) \ln \left( \frac{\alpha_n + (V^TED)_{mn}}{\alpha_n +V^T_{mn}}  \right)\right] \\
    &= \sum_{n=1}^N \sum_{m=1}^M \left[ V^T_{mn}\ln \left(V^T_{mn}\right) - V^T_{mn} \ln \left( (V^TED)_{mn} \right) - (\alpha_n + V^T_{mn}) \ln \left( \frac{\alpha_n + V^T_{mn}}{\alpha_n +(V^TED)_{mn}}  \right)\right] \\
    &= F(E; D, V).
\end{align*}

The update is then obtained by differentiating the auxiliary function, setting it equal to 0 to find the minimum and solving for $E^{(t+1)} = \arg\min_{E>0} G(E,E^t)$.

$$
E^{(t+1)} = E^{t} \cdot \frac{\left( V \left( \frac{V^T}{(V^TE^tD)}\right)D^T\right)}{\left( V\left( \frac{\alpha + V^T}{\alpha +(V^TE^tD)}\right)D^T\right)},
$$ 

For matrix-wise updates we here define the matrix $\alpha \in \mathbb{R}^+_{M \times N}$, obtained by matrix multiplying an $M\times 1$ column vector of ones with the $1\times N$ $\alpha$ vector defined previously.
\subsection{Update of D matrix}
Similarly we derive an auxiliary function for $D$ satisfying property (i) as before:
\begin{align*}
    G(D, D^t) = \sum_{n=1}^N \sum_{m=1}^M & \left[ V^T_{mn}\ln \left(V^T_{mn}\right) - V^T_{mn} \sum_{k=1}^K c_{k} \ln \left( \frac{(V^TE)_{mk}D_{kn} }{c_{k}} \right) +  \right. \\
    & \left. (\alpha_n + V^T_{mn}) \ln \left( \frac{\alpha_n + (V^TED^t)_{mn}}{\alpha_n +V^T_{mn}}  \right)+ \frac{\alpha_n + V^T_{mn} }{\alpha_n +(V^TED^t)_{mn} }((V^TED)_{mn} - (V^TED^t)_{mn})\right]
\end{align*}
Where $c_k = \frac{(V^TE)_{mk}D^t_{kn}}{(V^TED^t)_{mn}}$ and $\sum_{k=1}^K c_k = 1$.
We now show property (ii) of the auxiliary function is satisfied, such that $G(D,D) = F(D; E, V)$.

\begin{align*}
        G(D, D) &= \sum_{n=1}^N \sum_{m=1}^M \left[ V^T_{mn}\ln \left(V^T_{mn}\right) - V^T_{mn} \sum_{k=1}^K \frac{(V^TE)_{mk}D_{kn}}{(V^TED)_{mn}} \ln \left( \frac{(V^TE)_{mk}D_{kn} }{\frac{(V^TE)_{mk}D_{kn}}{(V^TED)_{mn}}} \right)  \right. \\
    & \qquad \qquad\qquad\left.+ (\alpha_n + V^T_{mn}) \ln \left( \frac{\alpha_n + (V^TED)_{mn}}{\alpha_n +V^T_{mn}}  \right) + \frac{\alpha_n + V^T_{mn} }{\alpha_n +(V^TED)_{mn} }((V^TED)_{mn} - (V^TED)_{mn})\right] \\
    &= \sum_{n=1}^N \sum_{m=1}^M \left[ V^T_{mn}\ln \left(V^T_{mn}\right) - V^T_{mn} \sum_{k=1}^K \frac{(V^TE)_{mk}D_{kn}}{(V^TED)_{mn}} \ln \left( (V^TED)_{mn} \right) \right. \\
    & \qquad \qquad\qquad\left. + (\alpha_n + V^T_{mn}) \ln \left( \frac{\alpha_n + (V^TED)_{mn}}{\alpha_n +V^T_{mn}}  \right)\right] \\
    &=  \sum_{n=1}^N \sum_{m=1}^M \left[ V^T_{mn}\ln \left(V^T_{mn}\right) - V^T_{mn}\ln \left( (V^TED)_{mn} \right) - (\alpha_n + V^T_{mn}) \ln \left( \frac{\alpha_n + V^T_{mn}}{\alpha_n +(V^TED)_{mn}}  \right)\right] \\
    &= F(D; E, V).
\end{align*} 
Again by differentiating $G(D,D^t)$, setting the derivative equal to zero and solving for $D$ we obtain $D^{(t+1)} = \arg \min_{D>0} G(D,D^t)$. Due to the relation between the decoder matrix $D$ in C-NMF and the weights matrix $W$ in NMF, we find that this update is equivalent to equation (13) in \citep{Pelizzola2023}; taking $D^T = W$ and $(V^TE)^T = H$ gives the equivalence.
 
As before we define the matrix $\alpha \in \mathbb{R}^+_{M \times N}$ by matrix multiplying an $M\times 1$ column vector of ones with the $1\times N$ $\alpha$ vector defined previously.
This gives the update
$$
D^{(t+1)} = D^{t}\cdot\frac{\left((V^TE)^T \left( \frac{V^T}{(V^TED^t)}\right)\right)}{\left( (V^TE)^T \left( \frac{\alpha + V^T}{\alpha +(V^TED^t)}\right)\right)}
$$

\section{Additional results on estimating topic domains in the newsgroups data} \label{app:D}
\begin{figure}[h!]
    \centering
    \includegraphics[width=\linewidth]{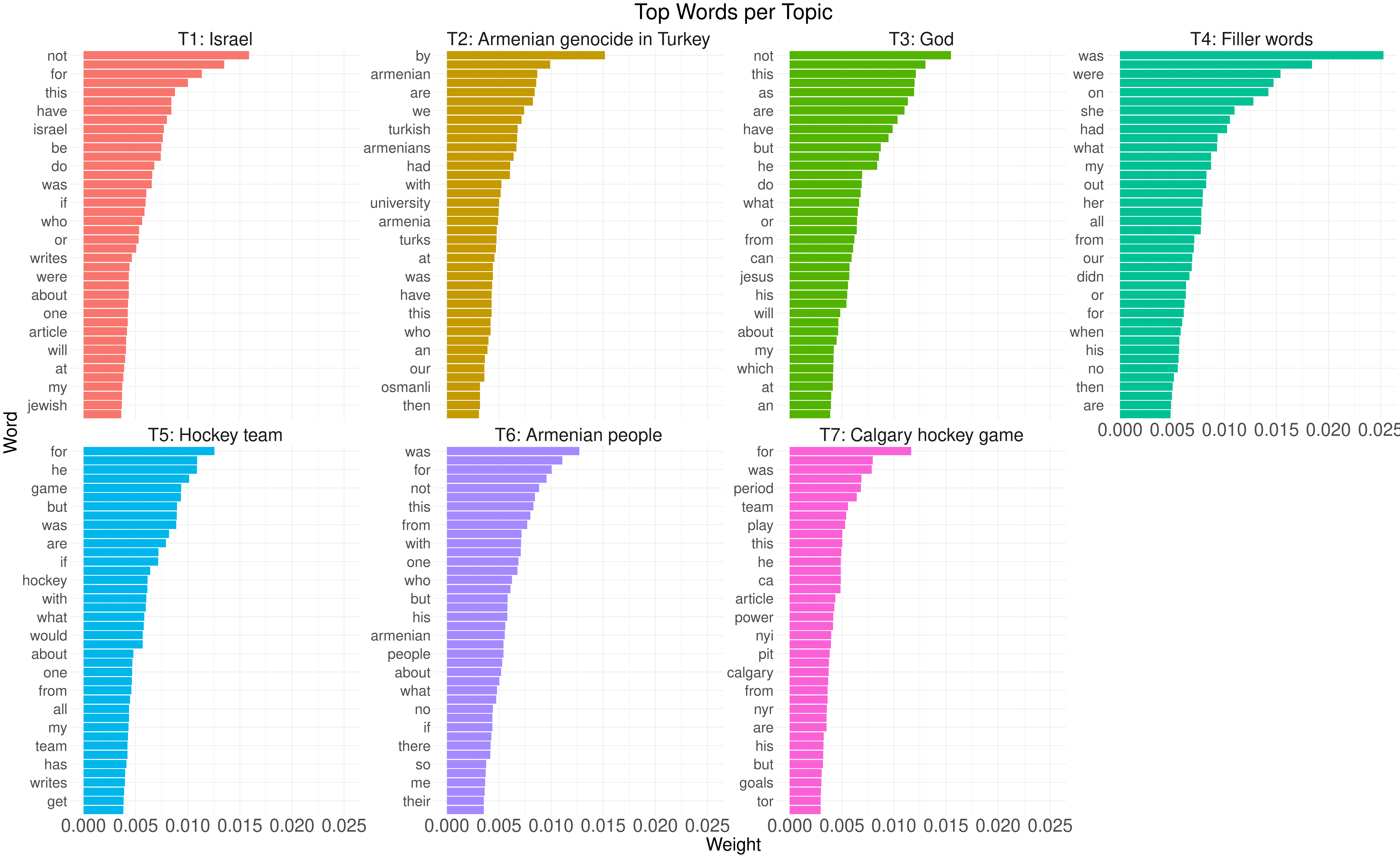}
    \caption{Top words for the topical domains estimated by  NMF/$\mathcal{C}$/TW$_{1.02}$/7 and their corresponding weight in the respective topic.}
    \label{fig:convextweedietopics}
\end{figure}

\begin{figure}[h!]
    \centering
    \includegraphics[width=\linewidth]{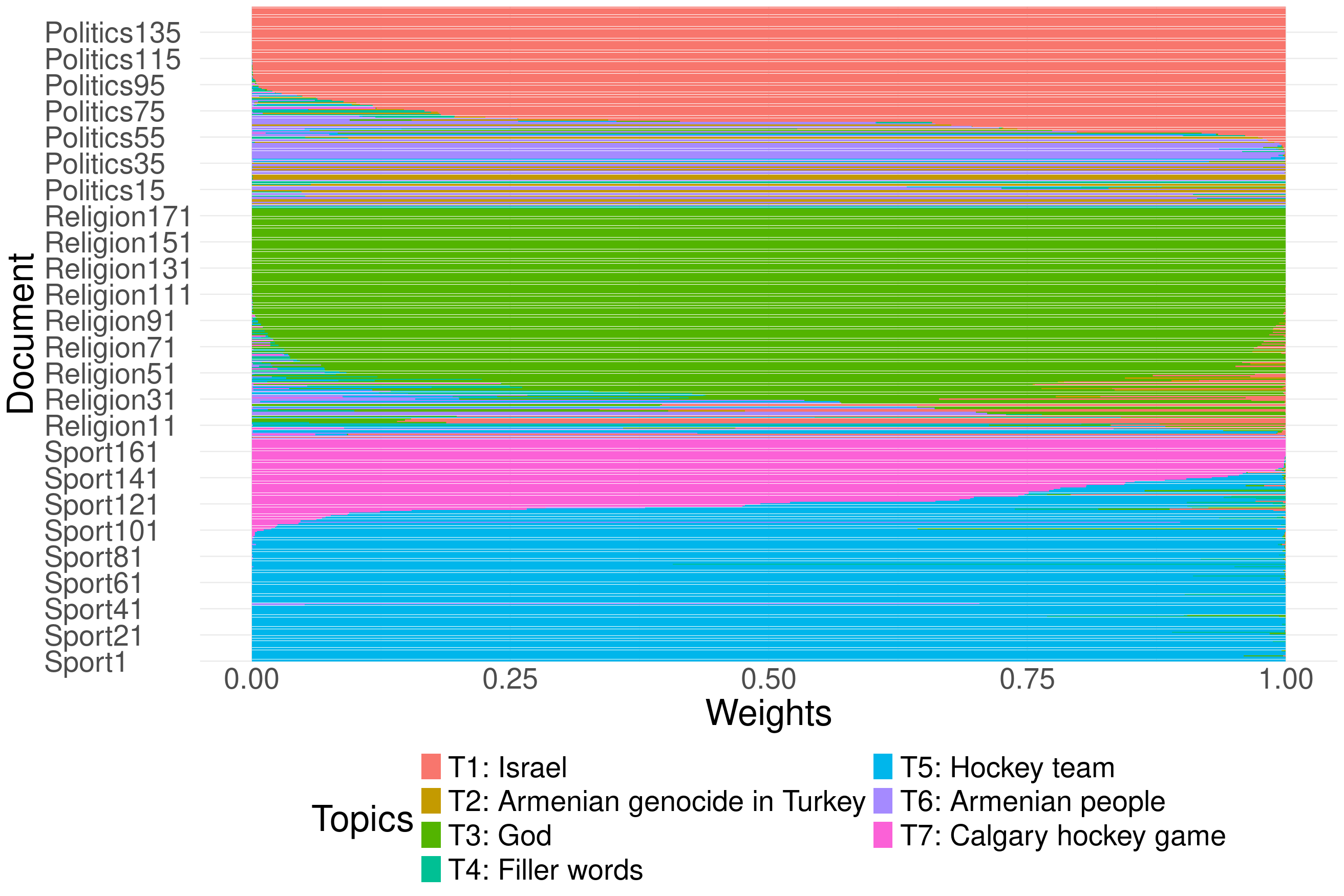}
    \caption{Weights (D matrix) for NMF/$\mathcal{C}$/TW$_{1.02}$/7. The weight of all topics for each document are represented by the different colours of the horizontal bars. Each bar correspond to a document. Documents are ordered by their original labels. }
    \label{fig:convextweedieweights}
\end{figure}

\newpage

\end{appendices}
\newpage

\bibliographystyle{apalike}
\bibliography{bibliography}

@book{dunn2018generalized,
  title={Generalized linear models with examples in R},
  author={Dunn, Peter K and Smyth, Gordon K and others},
  volume={53},
  year={2018},
  publisher={Springer}
}

@article{kendal2000characterization,
  title={Characterization of the frequency distribution for human hematogenous metastases: Evidence for clustering and a power variance function},
  author={Kendal, Wayne S and Lagerwaard, Frank J and Agboola, Olusegun},
  journal={Clinical \& Experimental Metastasis},
  volume={18},
  pages={219--229},
  year={2000},
  publisher={Springer}
}

@book{lange2020algorithms,
  title={Algorithms from the Book},
  author={Lange, Kenneth},
  year={2020},
  publisher={SIAM}
}

@book{jorgensen1997theory,
  title={The theory of dispersion models},
  author={J{\o}rgensen, Bent},
  year={1997},
  publisher={CRC Press}
}

@article{kendal2002spatial,
  title={{Spatial aggregation of the Colorado potato beetle described by an exponential dispersion model}},
  author={Kendal, Wayne S},
  journal={Ecological modelling},
  volume={151},
  number={2-3},
  pages={261--269},
  year={2002},
  publisher={Elsevier}
}

@article{enki2017taylor,
  title={Taylor's power law and the statistical modelling of infectious disease surveillance data},
  author={Enki, Doyo Gragn and Noufaily, Angela and Farrington, Paddy and Garthwaite, Paul and Andrews, Nick and Charlett, Andre},
  journal={Journal of the Royal Statistical Society Series A: Statistics in Society},
  volume={180},
  number={1},
  pages={45--72},
  year={2017},
  publisher={Oxford University Press}
}

@article{Laursen2022,
  title = {A Sampling Algorithm to Compute the Set of Feasible Solutions for NonNegative Matrix Factorization with an Arbitrary Rank},
  volume = {43},
  ISSN = {1095-7162},
  url = {http://dx.doi.org/10.1137/20M1378971},
  DOI = {10.1137/20m1378971},
  number = {1},
  journal = {SIAM Journal on Matrix Analysis and Applications},
  publisher = {Society for Industrial & Applied Mathematics (SIAM)},
  author = {Laursen,  Ragnhild and Hobolth,  Asger},
  year = {2022},
  month = feb,
  pages = {257–273}
}

@article{Pelizzola2023,
  doi = {10.1186/s12859-023-05304-1},
  url = {https://doi.org/10.1186/s12859-023-05304-1},
  year = {2023},
  month = may,
  publisher = {Springer Science and Business Media {LLC}},
  volume = {24},
  number = {1},
  author = {Marta Pelizzola and Ragnhild Laursen and Asger Hobolth},
  title = {Model selection and robust inference of mutational signatures using Negative Binomial non-negative matrix factorization},
  journal = {{BMC} Bioinformatics}
}

@article{alexandrov2013signatures,
  title={Signatures of mutational processes in human cancer},
  author={Alexandrov, Ludmil B and Nik-Zainal, Serena and Wedge, David C and Aparicio, Samuel AJR and Behjati, Sam and Biankin, Andrew V and Bignell, Graham R and Bolli, Niccolo and Borg, Ake and B{\o}rresen-Dale, Anne-Lise and others},
  journal={Nature},
  volume={500},
  number={7463},
  pages={415--421},
  year={2013},
  publisher={Nature Publishing Group}
}

@article{Gouvert2020,
    author = {Gouvert, Olivier and Oberlin, Thomas and Fevotte, Cedric},
    doi = {10.1109/LSP.2020.2991613},
    issn = {15582361},
    journal = {IEEE Signal Processing Letters},
    pages = {815--819},
    publisher = {Institute of Electrical and Electronics Engineers Inc.},
    title = {{Negative Binomial Matrix Factorization}},
    volume = {27},
    year = {2020}
}

@article{fevotte2009nonnegative,
  title={{Nonnegative matrix factorization with the Itakura-Saito divergence: With application to music analysis}},
  author={F{\'e}votte, C{\'e}dric and Bertin, Nancy and Durrieu, Jean-Louis},
  journal={Neural computation},
  volume={21},
  number={3},
  pages={793--830},
  year={2009},
  publisher={MIT Press}
}

@article{fevotte2011,
    author = {F{\'{e}}votte, C{\'{e}}dric and Idier, J{\'{e}}r{\^{o}}me},
    journal = {Neural Computation},
    doi = {10.1162/NECO_a_00168},
    eprint = {1010.1763},
    issn = {1530888X},
    number = {9},
    pages = {2421--2456},
    publisher = {MIT Press Journals},
    title = {Algorithms for nonnegative matrix factorization with the $\beta$-divergence},
    volume = {23},
    year = {2011}
}

@article{Lal2021,
    author = {Lal, Avantika and Liu, Keli and Tibshirani, Robert and Sidow, Arend and Ramazzotti, Daniele},
    doi = {10.1371/JOURNAL.PCBI.1009119},
    issn = {1553-7358},
    journal = {PLOS Computational Biology},
    month = {jun},
    number = {6},
    pages = {e1009119},
    publisher = {Public Library of Science},
    title = {{De novo mutational signature discovery in tumor genomes using SparseSignatures}},
    volume = {17},
    year = {2021}
}

@article{Lee1999,
    author = {Lee, Daniel D. and Seung, H. Sebastian},
    doi = {10.1038/44565},
    issn = {00280836},
    journal = {Nature},
    month = {oct},
    number = {6755},
    pages = {788--791},
    pmid = {10548103},
    publisher = {Nature Publishing Group},
    title = {{Learning the parts of objects by non-negative matrix factorization}},
    volume = {401},
    year = {1999}
}

@article{Lyu2020,
    author = {Lyu, Xinrui and Garret, Jean and R{\"{a}}tsch, Gunnar and Lehmann, Kjong Van},
    doi = {10.1093/BIOINFORMATICS/BTAA473},
    issn = {14602059},
    journal = {Bioinformatics},
    month = {jul},
    number = {Suppl{\_}1},
    pages = {i154--i160},
    pmid = {32657388},
    publisher = {Oxford University Press},
    title = {{Mutational signature learning with supervised negative binomial non-negative matrix factorization}},
    volume = {36},
    year = {2020}
}

@article{Campbell2020,
    author = {{Campbell et al.}},
    doi = {10.1038/s41586-020-1969-6},
    issn = {14764687},
    journal = {Nature},
    number = {7793},
    pages = {82--93},
    pmid = {32025007},
    publisher = {Nature Research},
    title = {{Pan-cancer analysis of whole genomes}},
    url = {https://doi.org/10.1038/s41586-020-1969-6},
    volume = {578},
    year = {2020}
}

@article{Alexandrov2020,
    author = {Alexandrov, Ludmil B. and Kim, Jaegil and Haradhvala, Nicholas J. and Huang, Mi Ni and {Tian Ng}, Alvin Wei and Wu, Yang and Boot, Arnoud and Covington, Kyle R. and Gordenin, Dmitry A. and Bergstrom, Erik N. and Islam, S. M. Ashiqul and Lopez-Bigas, Nuria and Klimczak, Leszek J. and McPherson, John R. and Morganella, Sandro and Sabarinathan, Radhakrishnan and Wheeler, David A. and Mustonen, Ville and Getz, Gad and Rozen, Steven G. and Stratton, Michael R.},
    doi = {10.1038/s41586-020-1943-3},
    issn = {1476-4687},
    journal = {Nature},
    month = {feb},
    number = {7793},
    pages = {94--101},
    publisher = {Nature Publishing Group},
    title = {{The repertoire of mutational signatures in human cancer}},
    volume = {578},
    year = {2020}
}

@article{Tate2019,
    author = {Tate, John G and Bamford, Sally and Jubb, Harry C and Sondka, Zbyslaw and Beare, David M and Bindal, Nidhi and Boutselakis, Harry and Cole, Charlotte G and Creatore, Celestino and Dawson, Elisabeth and Fish, Peter and Harsha, Bhavana and Hathaway, Charlie and Jupe, Steve C and Kok, Chai Yin and Noble, Kate and Ponting, Laura and Ramshaw, Christopher C and Rye, Claire E and Speedy, Helen E and Stefancsik, Ray and Thompson, Sam L and Wang, Shicai and Ward, Sari and Campbell, Peter J and Forbes, Simon A},
    doi = {10.1093/NAR/GKY1015},
    issn = {0305-1048},
    journal = {Nucleic Acids Research},
    month = {jan},
    number = {D1},
    pages = {D941--D947},
    publisher = {Oxford Academic},
    title = {{COSMIC: the Catalogue Of Somatic Mutations In Cancer}},
    volume = {47},
    year = {2019}
}

@article{Lindberg2019,
  doi = {10.1073/pnas.1909021116},
  url = {https://doi.org/10.1073/pnas.1909021116},
  year = {2019},
  month = sep,
  publisher = {Proceedings of the National Academy of Sciences},
  volume = {116},
  number = {41},
  pages = {20411--20417},
  author = {Markus Lindberg and Martin Bostr\"{o}m and Kerryn Elliott and Erik Larsson},
  title = {Intragenomic variability and extended sequence patterns in the mutational signature of ultraviolet light},
  journal = {Proceedings of the National Academy of Sciences}
}

@article{Caruso2017,
  title = {Niraparib in ovarian cancer: results to date and clinical potential},
  volume = {9},
  ISSN = {1758-8359},
  url = {http://dx.doi.org/10.1177/1758834017718775},
  DOI = {10.1177/1758834017718775},
  number = {9},
  journal = {Therapeutic Advances in Medical Oncology},
  publisher = {SAGE Publications},
  author = {Caruso,  Davide and Papa,  Anselmo and Tomao,  Silverio and Vici,  Patrizia and Panici,  Pierluigi Benedetti and Tomao,  Federica},
  year = {2017},
  month = jul,
  pages = {579–588}
}

@article{Zhang2021,
  title = {Genomic and evolutionary classification of lung cancer in never smokers},
  volume = {53},
  ISSN = {1546-1718},
  url = {http://dx.doi.org/10.1038/s41588-021-00920-0},
  DOI = {10.1038/s41588-021-00920-0},
  number = {9},
  journal = {Nature Genetics},
  publisher = {Springer Science and Business Media LLC},
  author = {Zhang,  Tongwu and Joubert,  Philippe and Ansari-Pour,  Naser and Zhao,  Wei and Hoang,  Phuc H. and Lokanga,  Rachel and Moye,  Aaron L. and Rosenbaum,  Jennifer and Gonzalez-Perez,  Abel and Martínez-Jiménez,  Francisco and Castro,  Andrea and Muscarella,  Lucia Anna and Hofman,  Paul and Consonni,  Dario and Pesatori,  Angela C. and Kebede,  Michael and Li,  Mengying and Gould Rothberg,  Bonnie E. and Peneva,  Iliana and Schabath,  Matthew B. and Poeta,  Maria Luana and Costantini,  Manuela and Hirsch,  Daniela and Heselmeyer-Haddad,  Kerstin and Hutchinson,  Amy and Olanich,  Mary and Lawrence,  Scott M. and Lenz,  Petra and Duggan,  Maire and Bhawsar,  Praphulla M. S. and Sang,  Jian and Kim,  Jung and Mendoza,  Laura and Saini,  Natalie and Klimczak,  Leszek J. and Islam,  S. M. Ashiqul and Otlu,  Burcak and Khandekar,  Azhar and Cole,  Nathan and Stewart,  Douglas R. and Choi,  Jiyeon and Brown,  Kevin M. and Caporaso,  Neil E. and Wilson,  Samuel H. and Pommier,  Yves and Lan,  Qing and Rothman,  Nathaniel and Almeida,  Jonas S. and Carter,  Hannah and Ried,  Thomas and Kim,  Carla F. and Lopez-Bigas,  Nuria and Garcia-Closas,  Montserrat and Shi,  Jianxin and Bossé,  Yohan and Zhu,  Bin and Gordenin,  Dmitry A. and Alexandrov,  Ludmil B. and Chanock,  Stephen J. and Wedge,  David C. and Landi,  Maria Teresa},
  year = {2021},
  month = sep,
  pages = {1348–1359}
}

@article{Egendal2025relation,
  title={{On the Relation Between Linear Autoencoders and Non-Negative Matrix Factorization for Mutational Signature Extraction}},
  author={Egendal, Ida and Br{\o}ndum, Rasmus Froberg and Pelizzola, Marta and Hobolth, Asger and B{\o}gsted, Martin},
  journal={Journal of Computational Biology},
  year={2025},
  publisher={Mary Ann Liebert, Inc., publishers 140 Huguenot Street, 3rd Floor New~…}
}

@article{Ding2010,
  title = {{Convex and Semi-Nonnegative Matrix Factorizations}},
  volume = {32},
  ISSN = {0162-8828},
  url = {http://dx.doi.org/10.1109/TPAMI.2008.277},
  DOI = {10.1109/tpami.2008.277},
  number = {1},
  journal = {IEEE Transactions on Pattern Analysis and Machine Intelligence},
  publisher = {Institute of Electrical and Electronics Engineers (IEEE)},
  author = {Ding,  C.H.Q. and Tao Li and Jordan,  M.I.},
  year = {2010},
  month = jan,
  pages = {45–55}
}

@article{yilmaz2012alpha,
  title={{Alpha/Beta Divergences and Tweedie Models}},
  author={Yilmaz, Y Kenan and Cemgil, A Taylan},
  journal={arXiv preprint arXiv:1209.4280},
  year={2012}
}

@article{Kendall1953,
  title = {{Stochastic Processes Occurring in the Theory of Queues and their Analysis by the Method of the Imbedded Markov Chain}},
  volume = {24},
  ISSN = {0003-4851},
  url = {http://dx.doi.org/10.1214/aoms/1177728975},
  DOI = {10.1214/aoms/1177728975},
  number = {3},
  journal = {The Annals of Mathematical Statistics},
  publisher = {Institute of Mathematical Statistics},
  author = {Kendall,  David G.},
  year = {1953},
  month = sep,
  pages = {338–354}
}

@article{LloydSmith2005,
  title = {Superspreading and the effect of individual variation on disease emergence},
  volume = {438},
  ISSN = {1476-4687},
  url = {http://dx.doi.org/10.1038/nature04153},
  DOI = {10.1038/nature04153},
  number = {7066},
  journal = {Nature},
  publisher = {Springer Science and Business Media LLC},
  author = {Lloyd-Smith,  J. O. and Schreiber,  S. J. and Kopp,  P. E. and Getz,  W. M.},
  year = {2005},
  month = nov,
  pages = {355–359}
}

@article{Albers2010,
  title = {The optimal choice of negative binomial charts for monitoring high-quality processes},
  volume = {140},
  ISSN = {0378-3758},
  url = {http://dx.doi.org/10.1016/j.jspi.2009.07.005},
  DOI = {10.1016/j.jspi.2009.07.005},
  number = {1},
  journal = {Journal of Statistical Planning and Inference},
  publisher = {Elsevier BV},
  author = {Albers,  Willem},
  year = {2010},
  month = jan,
  pages = {214–225}
}

@inproceedings{Lam2008,
  title = {Non-negative matrix factorization for images with Laplacian noise},
  url = {http://dx.doi.org/10.1109/APCCAS.2008.4746143},
  DOI = {10.1109/apccas.2008.4746143},
  booktitle = {APCCAS 2008 - 2008 IEEE Asia Pacific Conference on Circuits and Systems},
  publisher = {IEEE},
  author = {Lam,  Edmund Y.},
  year = {2008},
  month = nov,
  pages = {798–801}
}

@article{Kim2007,
    author = {Kim, Hyunsoo and Park, Haesun},
    title = {Sparse non-negative matrix factorizations via alternating non-negativity-constrained least squares for microarray data analysis},
    journal = {Bioinformatics},
    volume = {23},
    number = {12},
    pages = {1495-1502},
    year = {2007},
    month = {05},
    issn = {1367-4803},
    doi = {10.1093/bioinformatics/btm134},
}

@article{PascualMontano2006,
  title = {{Nonsmooth nonnegative matrix factorization (nsNMF)}},
  volume = {28},
  ISSN = {0162-8828},
  url = {http://dx.doi.org/10.1109/tpami.2006.60},
  DOI = {10.1109/tpami.2006.60},
  number = {3},
  journal = {IEEE Transactions on Pattern Analysis and Machine Intelligence},
  publisher = {Institute of Electrical and Electronics Engineers (IEEE)},
  author = {Pascual-Montano,  A. and Carazo,  J.M. and Kochi,  K. and Lehmann,  D. and Pascual-Marqui,  R.D.},
  year = {2006},
  month = mar,
  pages = {403–415}
}

@article{Qian2024,
  title = {{scRNMF: An imputation method for single-cell RNA-seq data by robust and non-negative matrix factorization}},
  volume = {20},
  ISSN = {1553-7358},
  url = {http://dx.doi.org/10.1371/journal.pcbi.1012339},
  DOI = {10.1371/journal.pcbi.1012339},
  number = {8},
  journal = {PLOS Computational Biology},
  publisher = {Public Library of Science (PLoS)},
  author = {Qian,  Yuqing and Zou,  Quan and Zhao,  Mengyuan and Liu,  Yi and Guo,  Fei and Ding,  Yijie},
  editor = {Nie,  Qing},
  year = {2024},
  month = aug,
  pages = {e1012339}
}

@article{Mohammadiha2013,
  author={Mohammadiha, Nasser and Smaragdis, Paris and Leijon, Arne},
  journal={IEEE Transactions on Audio, Speech, and Language Processing}, 
  title={{Supervised and Unsupervised Speech Enhancement Using Nonnegative Matrix Factorization}}, 
  year={2013},
  volume={21},
  number={10},
  pages={2140-2151},
  doi={10.1109/TASL.2013.2270369}}

@article{sidorov2014soft,
  title={Soft similarity and soft cosine measure: Similarity of features in vector space model},
  author={Sidorov, Grigori and Gelbukh, Alexander and G{\'o}mez-Adorno, Helena and Pinto, David},
  journal={Computaci{\'o}n y Sistemas},
  volume={18},
  number={3},
  pages={491--504},
  year={2014},
  publisher={Instituto Polit{\'e}cnico Nacional, Centro de Investigaci{\'o}n en Computaci{\'o}n}
}

@article{Leplat2020,
  author={Leplat, Valentin and Gillis, Nicolas and Ang, Andersen M.S.},
  journal={IEEE Transactions on Signal Processing}, 
  title={{Blind Audio Source Separation With Minimum-Volume Beta-Divergence NMF}}, 
  year={2020},
  volume={68},
  number={},
  pages={3400-3410},
  doi={10.1109/TSP.2020.2991801}}

@article{Matsumoto2019,
    author = {Matsumoto, Hirotaka and Hayashi, Tetsutaro and Ozaki, Haruka and Tsuyuzaki, Koki and Umeda, Mana and Iida, Tsuyoshi and Nakamura, Masaya and Okano, Hideyuki and Nikaido, Itoshi},
    title = {{An NMF-based approach to discover overlooked differentially expressed gene regions from single-cell RNA-seq data}},
    journal = {NAR Genomics and Bioinformatics},
    volume = {2},
    number = {1},
    pages = {lqz020},
    year = {2019},
    month = {12},
    issn = {2631-9268},
    doi = {10.1093/nargab/lqz020},
}

@article{Kalcheva2020,
  title = {{Laplace Naive Bayes classifier in the classification of text in machine learning}},
  url = {http://dx.doi.org/10.1109/BIA50171.2020.9244491},
  DOI = {10.1109/bia50171.2020.9244491},
  journal = {2020 International Conference on Biomedical Innovations and Applications (BIA)},
  publisher = {IEEE},
  author = {Kalcheva,  Neli and Nikolov,  Nedyalko},
  year = {2020},
  month = sep,
  pages = {17–19}
}

@misc{Rennie2007,
  title = {20 newsgroups data set},
  howpublished = {\url{http://qwone.com/~jason/20Newsgroups}},
  author = {Rennie, J},
  year = {2007},
}

@article{onan2016,
  title={{LDA-based topic modelling in text sentiment classification: An empirical analysis.}},
  author={Onan, Aytug and Korukoglu, Serdar and Bulut, Hasan},
  journal={Int. J. Comput. Linguistics Appl.},
  volume={7},
  number={1},
  pages={101--119},
  year={2016}
}

@article{Lang1995,
    author = {Ken Lang},
    title = {Newsweeder: Learning to filter netnews},
    year = {1995},
    journal = {Proceedings of the Twelfth International Conference on Machine Learning},
    pages = {331-339}
}

@article{Ozer2022,
  title = {Investigating Nonnegative Autoencoders for Efficient Audio Decomposition},
  url = {http://dx.doi.org/10.23919/EUSIPCO55093.2022.9909787},
  DOI = {10.23919/eusipco55093.2022.9909787},
  journal = {2022 30th European Signal Processing Conference (EUSIPCO)},
  publisher = {IEEE},
  author = {Ozer,  Yigitcan and Hansen,  Jonathan and Zunner,  Tim and Muller,  Meinard},
  year = {2022},
  month = aug,
  pages = {254–258}
}

@article{Smaragdis2017,
  title = {A neural network alternative to non-negative audio models},
  url = {http://dx.doi.org/10.1109/ICASSP.2017.7952123},
  DOI = {10.1109/icassp.2017.7952123},
  journal = {2017 IEEE International Conference on Acoustics,  Speech and Signal Processing (ICASSP)},
  publisher = {IEEE},
  author = {Smaragdis,  Paris and Venkataramani,  Shrikant},
  year = {2017},
  month = mar,
  pages = {86–90}
}

@article{Pancotti2024,
  title = {{MUSE-XAE: MUtational Signature Extraction with eXplainable AutoEncoder enhances tumour types classification}},
  volume = {40},
  ISSN = {1367-4811},
  url = {http://dx.doi.org/10.1093/bioinformatics/btae320},
  DOI = {10.1093/bioinformatics/btae320},
  number = {5},
  journal = {Bioinformatics},
  publisher = {Oxford University Press (OUP)},
  author = {Pancotti,  Corrado and Rollo,  Cesare and Codicè,  Francesco and Birolo,  Giovanni and Fariselli,  Piero and Sanavia,  Tiziana},
  editor = {Kelso,  Janet},
  year = {2024},
  month = may 
}

@article{Pei2020,
  doi = {10.1038/s41388-020-1343-z},
  url = {https://doi.org/10.1038/s41388-020-1343-z},
  year = {2020},
  month = jun,
  publisher = {Springer Science and Business Media {LLC}},
  volume = {39},
  number = {27},
  pages = {5031--5041},
  author = {Guangsheng Pei and Ruifeng Hu and Yulin Dai and Zhongming Zhao and Peilin Jia},
  title = {Decoding whole-genome mutational signatures in 37 human pan-cancers by denoising sparse autoencoder neural network},
  journal = {Oncogene}
}

@article{Carbonetto2025,
    author = {Peter Carbonetto and Abhishek Sarkar and Zihao Wang and Matthew Stephens},
    title = {{Non-negative Matrix Factorization
Algorithms Generally Improve Topic Model
Fits}},
    journal = {arXiv},
    year = {2025} 
}

@article{Brunet2004,
  title = {Metagenes and molecular pattern discovery using matrix factorization},
  volume = {101},
  ISSN = {1091-6490},
  url = {http://dx.doi.org/10.1073/pnas.0308531101},
  DOI = {10.1073/pnas.0308531101},
  number = {12},
  journal = {Proceedings of the National Academy of Sciences},
  publisher = {Proceedings of the National Academy of Sciences},
  author = {Brunet,  Jean-Philippe and Tamayo,  Pablo and Golub,  Todd R. and Mesirov,  Jill P.},
  year = {2004},
  month = mar,
  pages = {4164–4169}
}

@article{Hobolth2019,
  title = {A Unifying Framework and Comparison of Algorithms for Non‐negative Matrix Factorisation},
  volume = {88},
  ISSN = {1751-5823},
  url = {http://dx.doi.org/10.1111/insr.12331},
  DOI = {10.1111/insr.12331},
  number = {1},
  journal = {International Statistical Review},
  publisher = {Wiley},
  author = {Hobolth,  Asger and Guo,  Qianyun and Kousholt,  Astrid and Jensen,  Jens Ledet},
  year = {2019},
  month = jul,
  pages = {29–53}
}

@article{Lange2000,
  title = {Optimization Transfer Using Surrogate Objective Functions},
  volume = {9},
  ISSN = {1537-2715},
  url = {http://dx.doi.org/10.1080/10618600.2000.10474858},
  DOI = {10.1080/10618600.2000.10474858},
  number = {1},
  journal = {Journal of Computational and Graphical Statistics},
  publisher = {Informa UK Limited},
  author = {Lange,  Kenneth and Hunter,  David R. and Yang,  Ilsoon},
  year = {2000},
  month = mar,
  pages = {1–20}
}

\end{document}